\documentclass[final]{cvpr}

\usepackage{times}
\usepackage{epsfig}
\usepackage{graphicx}
\usepackage{amsmath}
\usepackage{amssymb}
\usepackage{multirow}
\usepackage{array}
\usepackage{algorithm, algpseudocode}
\usepackage{color}
\usepackage{subfigure}

\let\oldReturn\Return
\renewcommand{\Return}{\State\oldReturn}

\usepackage[pagebackref=true,breaklinks=true,colorlinks,bookmarks=false]{hyperref}

\def\cvprPaperID{3644} 
\def\confYear{CVPR 2021}

\begin{document}

\title{Improving Transferability of Adversarial Patches on Face Recognition with Generative Models}

\newcommand*\samethanks[1][\value{footnote}]{\footnotemark[#1]}
\author{Zihao Xiao$^1$\thanks{Equal contributions.}~\thanks{Corresponding authors.}\quad Xianfeng Gao$^{1,4}$\samethanks[1]\quad Chilin Fu$^2$ \quad Yinpeng Dong$^{1,3}$ \quad Wei Gao$^{5}$\thanks{Work done as an intern at RealAI.} \\
Xiaolu Zhang$^{2}$ \quad Jun Zhou$^2$ \quad Jun Zhu$^{3}$\samethanks[2]\\
$^1$~RealAI\quad $^2$~Ant Financial \quad
$^3$ Tsinghua University \\
$^4$ Beijing Institute of Technology  \quad $^5$ Nanyang Technological University\\
{\tt\small zihao.xiao@realai.ai, ggxxff@bit.edu.cn, chilin.fcl@antgroup.com, dyp17@mails.tsinghua.edu.cn}\\
{\tt\small gaow0007@ntu.edu.sg, \{yueyin.zxl, jun.zhoujun\}@antfin.com, dcszj@tsinghua.edu.cn}
}

\maketitle
\pagestyle{empty}  
\thispagestyle{empty} 

\begin{abstract}
Face recognition is greatly improved by deep convolutional neural networks (CNNs). Recently, these face recognition models have been used for identity authentication in security sensitive applications. However, deep CNNs are vulnerable to adversarial patches, which are physically realizable and stealthy, raising new security concerns on the real-world applications of these models. In this paper, we evaluate the robustness of face recognition models using adversarial patches based on transferability, where the attacker has limited accessibility to the target models. First, we extend the existing transfer-based attack techniques to generate transferable adversarial patches. However, we observe that the transferability is sensitive to initialization and degrades when the perturbation magnitude is large, indicating the overfitting to the substitute models. Second, we propose to regularize the adversarial patches on the low dimensional data manifold. The manifold is represented by generative models pre-trained on legitimate human face images. Using face-like features as adversarial perturbations through optimization on the manifold, we show that the gaps between the responses of substitute models and the target models dramatically decrease, exhibiting a better transferability. Extensive digital world experiments are conducted to demonstrate the superiority of the proposed method in the black-box setting. We apply the proposed method in the physical world as well.
\end{abstract}

\section{Introduction}

\begin{figure}
    \centering
    \includegraphics[width=8cm]{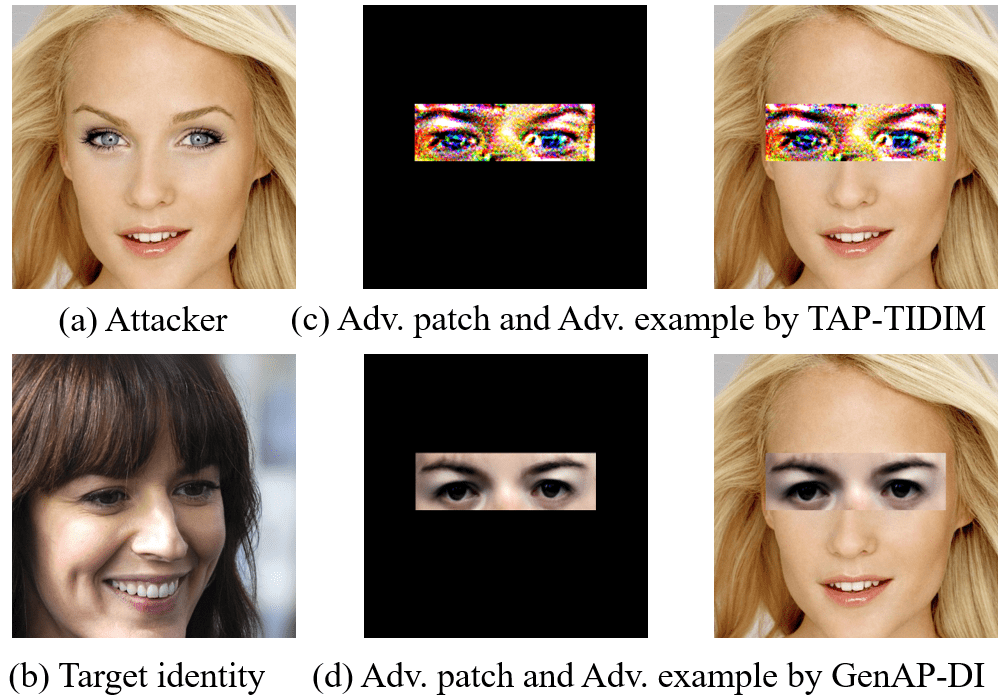}
    \caption{Demonstration of adversarial patches against face recognition models. \textbf{(a)} The attacker who wants to impersonate the target identity. \textbf{(b)} An image of the target identity. \textbf{(c)} The adversarial patch and the corresponding adversarial example generated by the TAP-TIDIM algorithm. \textbf{(d)} The adversarial patch and the corresponding adversarial example generated by the proposed GenAP-DI algorithm. The proposed GenAP algorithms use face-like features as perturbations to improve the transferability.}
    \label{fig:demo}
\end{figure}

Deep convolutional neural networks (CNNs) have led to substantial performance improvements on many computer vision tasks. As an important task, face recognition is also greatly facilitated by deep CNNs~\cite{schroff2015facenet,wang2018cosface,deng2019arcface}. Due to their excellent recognition performance, deep face recognition models have been used for identity authentication in security-sensitive applications, \eg, finance/payment, public access, face unlock on smart phones, \etc.

However, deep CNNs are shown to be vulnerable to adversarial examples at test time~\cite{szegedy2013intriguing,dong2020benchmarking}.
Adversarial examples are elaborately perturbed images that can fool models to make wrong predictions. Early adversarial examples on deep CNNs are indistinguishable from legitimate ones for human observers by slightly perturbing every pixel in an image. Later,~\cite{sharif2016accessorize} proposes adversarial patches, which only perturb a small cluster of pixels. Several works have shown that the adversarial patches can be made into physical objects to fool deep CNNs in the wild. For example, ~\cite{evtimov2017robust,song2018physical,wu2019making} use adversarial stickers or T-shirts to fool special purpose object detectors.~\cite{sharif2016accessorize} proposes an adversarial eyeglass frame to impersonate another identity against face recognition models. These works show that adversarial patches are physically realizable and stealthy. Using the adversarial patches in the physical world, the attacker can fool a recognition model without accessing the digital input to it, making them an emerging threat to deep learning applications, especially to face recognition systems in security-sensitive scenarios.

Previous works on adversarial patches are developed under the white-box setting~\cite{evtimov2017robust,song2018physical,brown2017adversarial,sharif2016accessorize}, where the attacker knows the parameters of the target model, or under the query-based setting~\cite{sharif2016accessorize,yang2020patchattack}, where the attacker can make many queries against the target model. But for a black-box model deployed in the wild, both the white-box information and the excessive queries are not easily attainable. In this paper, we focus on evaluating the robustness of face recognition models under the query-free black-box setting, which is a more severe and realistic threat model.

Under the query-free black-box setting, the adversarial attacks based on transferability are widely used. Transfer-based attacks~\cite{goodfellow2014explaining}
leverage that the adversarial examples for the white-box substitute models are also effective at the black-box target models.
Specifically, most adversarial algorithms perform optimization on an adversarial objective specified by the substitute models as a surrogate, to approximate the true (but unknown) adversarial objective on the black-box target models.
Existing techniques on improving the transferability of adversarial examples focus on using advanced non-convex optimization~\cite{dong2018boosting}, data  augmentations~\cite{xie2019improving,dong2019evading}, \etc.
These techniques are originally proposed to generate $\mathcal{L}_{p}$-norm $(p>0)$ constrained adversarial examples, and we show that they can be extended to improve the transferability of adversarial patches as well. 

However, even though these techniques are extended and applied in the patch setting, we still observe it easy for the optimization to be trapped into local optima with unsatisfactory transferability.  First, the transferability is sensitive to initialization of the algorithms. Second, if the perturbation magnitude increases, the transferability first rises and then falls, exhibiting an overfitting phenomenon. The difficulties in escaping solutions of unsatisfactory transferability indicate that the optimization is prone to overfitting the substitute model and new regularization methods are required.

We propose to regularize the adversarial patch by optimizing it on a low-dimensional manifold. Specifically, the manifold is represented by a generative model and the optimization is conducted in its latent space.
The generative model is pre-trained on legitimate human face data and can generate diverse and unseen human face images by manipulating the latent vectors to assemble different face features. By optimizing the adversarial objective on this latent space, the adversarial perturbations resemble human face features (see Fig.~\ref{fig:demo}, (d)), on which the predictions of the white-box substitute and the black-box target model are more related. Consequently, the overfitting problem is relieved and the transferability is improved.

Extensive experiments are conducted to show the superiority of the proposed method for black-box attacks on face recognition. 
We show its effectiveness in the physical world as well. Finally, we extend the proposed method to other tasks, \eg, image classification.

\section{Related work}

\subsection{Adversarial patches}
Most existing works on adversarial patches are designed for the white-box setting~\cite{sharif2016accessorize,evtimov2017robust,song2018physical,brown2017adversarial,wu2019making} or the query-based black-box setting~\cite{sharif2016accessorize,yang2020patchattack}. This paper focuses on the query-free black-box setting, a realistic assumption on the adversary's knowledge on the target models deployed in the wild~\cite{dong2018boosting}. Although some works demonstrate results on query-free attacks~\cite{brown2017adversarial,wu2019making}, their methods are not optimized for this setting and not optimal.

\subsection{Transferable adversarial examples}
There are many works proposed for improving the transferability of adversarial examples, and most of them are developed under the $\mathcal{L}_p$-norm constrained setting~\cite{dong2018boosting,xie2019improving,dong2019evading}. In contrast, we focus on adversarial patches, a different condition on the adversary's capacity to perturb the visual inputs. Adversarial patches are physically realizable and stealthy, posing threats to target models deployed in the wild. In this paper, we show that while many methods proposed for the $\mathcal{L}_p$-norm constrained setting are useful for the patch setting, they are still prone to overfitting the substitute models and new regularization techniques are required.

\subsection{Generative modeling for adversarial examples}
Researchers have discovered that using generative models to generate adversarial examples has advantages. For example, efficient attack algorithms are proposed for white-box attacks~\cite{xiao2018generating} and  query-based attacks~\cite{tu2019autozoom,zhao2017generating}. Emerging threat models are studied using generative models as well, \eg., unrestricted adversarial examples~\cite{song2018constructing} and semantic adversarial examples~\cite{qiu2020semanticadv}. Unrestricted adversarial examples are closely related to adversarial patches, but~\cite{song2018constructing} does not show an improvement of transferability. Although SemanticAdv~\cite{qiu2020semanticadv} claims an improvement of transferability in their setting\footnote{They consider semantic perturbations.}, we show that it is sub-optimal in the patch setting. Our work shows how to adequately use generative models to improve the transferability of adversarial patches.

\section{Methodology}
This section introduces our method of generating adversarial patches on face recognition models with generative models. Sec.~\ref{sec:threat model} introduces the attack setting. Sec.~\ref{sec:pixel space optimization} extends the existing transfer-based attack methods from the $\mathcal{L}_p$-norm constrained setting to the patch setting, and show their problems. Sec.~\ref{sec:latent space optimization} elaborates the proposed method.

\subsection{Attack setting}\label{sec:threat model}
Face recognition usually includes face verification and face identification. The former identifies whether two face images belong to the same identity, while the latter classifies an image to a specific identity. For face verification, the similarity between two faces are compared with a threshold to give the prediction. For face identification, the similarity between a face image and those of a gallery set of face images is compared, and the input image is recognized as the identity whose representation is most similar to its.

Let $f(\mathbf{x}):\mathcal{X}\rightarrow \mathbb{R}^d$ denote a face recognition model that extracts a normalized feature representation vector for an input image $\mathbf{x}\in \mathcal{X}\subset \mathbb{R}^n$. Given a pair of face images $\{\mathbf{x}_s, \mathbf{x}_t\}$, the face recognition model estimates the similarity between the two faces by calculating the distance between the feature vectors extracted from the two images
\begin{align}
    \mathcal{D}_f(\mathbf{x}_s, \mathbf{x}_t) = ||f(\mathbf{x}_s) - f(\mathbf{x}_t)||_2^2.\label{equ:similarity}
\end{align}
And face verification and identification methods are done based on this similarity score $\mathcal{D}_f$ or its simple variants.

An adversary has generally two goals against the face recognition models --- dodging and impersonation.
Dodging attack aims to generate an adversarial face image that is recognized wrongly, which can be utilized to protect privacy against excessive surveillance. For face verification, the adversary can modify one image from a pair of images belonging to the same identity, to make the model recognize them as different identities. For face identification, the adversary generates an adversarial face image such that it is recognized as any other false identity.

Impersonation attack corresponds to generating an adversarial face image that is recognized as an adversary-specified target identity, which could be used to evade the face authentication systems. For face verification, the attacker aims to find an adversarial image that is recognized as the same identity of another image, while the original images are from different identities. For face identification, the generated adversarial image is expected to be classified as a specific identity.

\subsection{Transferable adversarial patch}\label{sec:pixel space optimization}
In the query-free black-box setting, the detailed information of the target model is unknown and an excessive amount of queries are not allowed. The adversarial attacks based on transferability~\cite{dong2018boosting,xie2019improving} show that, the adversarial examples for some white-box substitute model $g$ can remain adversarial for the black-box target model $f$. We focus on generating transferable adversarial patches (TAPs).

Suppose $g$ is a white-box face recognition model that is accessible to the attacker, and it can also define a similarity score $\mathcal{D}_g(\mathbf{x}_s, \mathbf{x}_t)$ for face recognition, similar to Eq.~\eqref{equ:similarity}. An adversary solves the following optimization problem to generate the adversarial patch on the substitute model~\cite{sharif2016accessorize}:
\begin{align}
    \max_{\mathbf{x}}  ~~&\mathcal{L}_g(\mathbf{x}, \mathbf{x}_t),\nonumber\\
    \text{s.t.  } \mathbf{x} \odot (1 - \mathbf{M}) &= \mathbf{x}_s \odot (1 - \mathbf{M}),\label{equ:black-box objective}
\end{align}
where $\mathcal{L}_g$ is a differentiable adversarial objective, $\odot$ is the element-wise product, and $\mathbf{M}\in \{0, 1\}^n$ is a binary mask. The constrain emphasizes that only the pixels whose corresponding elements in $\mathbf{M}$ are $1$ can be perturbed. Fig.~\ref{fig:demo} demonstrates how the masks $\mathbf{M}$ control the regions of the adversarial patches. We use $\mathcal{L}_g = \mathcal{D}_g$ for dodging attack and $\mathcal{L}_g = - \mathcal{D}_g$ for impersonation attack, respectively. In this paper, we fix the adversarial loss to fairly compare different techniques operated on the input $\mathbf{x}$.

Existing works on improving the transferability of adversarial examples focus on the $\mathcal{L}_p$-norm constrained setting. We can extend them to the patch setting. The vanilla algorithm is to use the momentum iterative method (MIM)~\cite{dong2018boosting} to solve the optimization problem~\eqref{equ:black-box objective}. We denote this algorithm as TAP-MIM. Advanced techniques to improve the transferability can be applied, \eg the data augmentations in TI-DIM~\cite{xie2019improving,dong2019efficient}. The overall algorithm is depicted in Alg.~\ref{alg:TAP} and is denoted as TAP-TIDIM. In the experiment (Sec.~\ref{sec:experimental results}), we show that TAP-TIDIM outperforms TAP-MIM, indicating that methods proposed for the $\mathcal{L}_p$-norm constrained setting might also be useful for the patch setting. Note that the TAP-TIDIM algorithm is similar with using the EoT technique~\cite{athalye2018synthesizing} to generate universal and physical-world adversarial patches in the white-box setting~\cite{brown2017adversarial,evtimov2017robust,song2018physical}

\begin{algorithm}[t]
\small
\caption{Transferable Adversarial Patch: TAP-TIDIM}
\label{alg:TAP}
\begin{algorithmic}[1]
\Require The adversarial objective function $\mathcal{L}_g$; a real face image $\mathbf{x}_s$ of the attacker; a  real face images $\mathbf{x}_t$ of the target identity; a binary mask matrix $\mathbf{M}$.
\Require A set of transformations $\mathcal{T}$; the size of perturbation $\epsilon$; learning rate $\alpha$; iterations $N$; decay factor $\mu$.
\Ensure
An adversarial image $\mathbf{x}^*$ by solving Eq.~\eqref{equ:black-box objective}.
\State $\mathbf{g}_0 = 0$;
\State $\bar{\mathbf{x}}_0 = \mathbf{x}_s$;\label{equ:initialization}
\For {$n = 0$ to $N-1$}
\State Sample a transformation $T$ from $\mathcal{T}$;
\State Blend the adversarial patch to $\mathbf{x}_s$
\begin{align}
\mathbf{x}_{n}^* &= \mathbf{x}_s \odot (1 - \mathbf{M}) + \bar{\mathbf{x}}_{n} \odot \mathbf{M};\nonumber
\end{align}
\State Input $T(\mathbf{x}_n^*)$ and obtain the loss $\mathcal{L}_g(T(\mathbf{x}_n^*), \mathbf{x}_t)$
\State Obtain the gradient $\nabla_{\mathbf{x}=\bar{\mathbf{x}}_n}\mathcal{L}_{g}(T(\mathbf{x}_n^*))$;
\State Convolve the gradient as in~\cite{dong2019evading}
\begin{equation}
\mathbf{W} * \nabla_{\mathbf{x}}\mathcal{L}_{g}(T(\mathbf{x}_{n}^*)),\nonumber
\end{equation}
where $\mathbf{W}$ is the Gaussian kernel and $*$ is convolution;
\State Update $\mathbf{g}_{t+1}$ as in~\cite{dong2018boosting}
\vspace{-2ex}
\begin{equation}
\mathbf{g}_{n+1} = \mu \cdot \mathbf{g}_{n} + \frac{\mathbf{W}*\nabla_{\mathbf{x}}\mathcal{L}_{g}(T(\mathbf{x}_{n}^*))}{\|\mathbf{W}*\nabla_{\mathbf{x}}\mathcal{L}_{g}(T(\mathbf{x}_{n}^*))\|_1};\nonumber
\end{equation}
\vspace{-3ex}
\State Update $\bar{\mathbf{x}}_{n+1}$ by applying the sign gradient as
\vspace{-1.5ex}
\begin{align}
\bar{\mathbf{x}}_{n+1} &= \mathrm{Clip}_{[\mathbf{x}^*_0-\epsilon, \mathbf{x}^*_0+\epsilon]}\big( \bar{\mathbf{x}}_{n} - \alpha\cdot\mathrm{sign}(\mathbf{g}_{n+1})\big);\nonumber
\end{align}
\EndFor
\Return $\mathbf{x}^* = \mathbf{x}_s \odot (1 - \mathbf{M}) + \bar{\mathbf{x}}_{N}\odot \mathbf{M}$.
\end{algorithmic}
\end{algorithm}

However,  even for the more advanced TAP-TIDIM algorithm, it is still difficult for the optimization to escape local optima with unsatisfactory transferability. Specifically, we observe the following two phenomena in our ablation studies (the details are in Sec.~\ref{sec:ablation study-TAP}):

\textbullet~~\textbf{The transferability is sensitive to the initialization of the optimization.} Note that TAP-TIDIM uses the face image of the attacker to initialize the patch, \ie, $\bar{\mathbf{x}}_0 = \mathbf{x}_s$ (see line~\ref{equ:initialization} of Alg.~\ref{alg:TAP}). For the impersonation attack, a simple modification is to use the face image of the target identity to initialize the patch, \ie, $\bar{\mathbf{x}}_0 = \mathbf{x}_t$. The modified algorithm is denoted as TAP-TIDIMv2. Experiments show that, TAP-TIDIMv2 finds solutions with significantly higher transferability than TAP-TIDIM by simply changing the initialization step (see Tab.~\ref{tab:impersonate-verification-eyeglass}).

\textbullet~~\textbf{The transferability degrades when the search space is large.} Specifically, we apply an additional $\mathcal{L}_{\infty}$-norm constrain on the optimization problem~(\ref{equ:black-box objective}) to control the size of the search space, i.e., $|(\mathbf{x} - \mathbf{x}_s)\odot \mathbf{M}|_{\infty}\leq \epsilon$. The $\mathcal{L}_{\infty}$-norm constrain bounds the maximum allowable perturbation magnitude~\cite{goodfellow2014explaining}. Our ablation studies show that, when $\epsilon$ increases, the transferability first rises and then falls (see Fig.~\ref{fig:epislon}).

The aforementioned two phenomena 
are indicators that the optimization problem~\eqref{equ:black-box objective} has many local optima of unsatisfactory transferability and the adversarial patches are overfitting the substitute model. It is hard to escape from these solutions even though many existing regularization techniques~\cite{dong2018boosting,xie2019improving,dong2019evading} have been applied in TAP-TIDIM.
Therefore, we resort to new regularization methods for the patch setting in the following section.

\subsection{Generative adversarial patch}\label{sec:latent space optimization}
We propose to optimize the adversarial patch on a low-dimensional manifold as a regularization to escape from the local optima of unsatisfactory transferability in the optimization problem~\eqref{equ:black-box objective}. The manifold poses a specific structure on the optimization dynamics. We consider a good manifold should have the following properties:

\noindent \textbf{1. Sufficient capacity.} The manifold should have a sufficient capacity so that the white-box attack on the substitute model is successful.

\noindent \textbf{2. Well regularized.} The manifold should be well regularized so that the responses of the substitute models and the target models are effectively related to avoid overfitting the substitute models.

To balance the demands for capacity and regularity, we use the manifold learnt by a generative model, where the generative model is pre-trained on natural human face data. Specifically, let $h(\mathbf{s}):\mathcal{S}\rightarrow\mathbb{R}^n$ denote a pre-trained generative model and $\mathcal{S}$ is its latent space.
The generative model can generate diverse and unseen human faces by manipulating the latent vector to assemble different face features, \eg, the color of eyeballs, the thickness of eyebrows, \etc. We propose to use the generative model to generate the adversarial patch, and optimize the patch through the latent vector. The optimization problem~\eqref{equ:black-box objective} becomes:
\begin{align}
    \max_{\mathbf{s}\in\mathcal{S}}  ~~&\mathcal{L}_g(\mathbf{x}, \mathbf{x}_t),\nonumber\\
    \text{s.t.  } \mathbf{x} \odot (1 - \mathbf{M}) &= \mathbf{x}_s \odot (1 - \mathbf{M}),\nonumber\\
    \mathbf{x}\odot \mathbf{M} &= h(\mathbf{s})\odot \mathbf{M} \label{equ:black-box objective - latent space}
\end{align}
where the second constrain restricts the adversarial patch on the low-dimensional manifold represented by the generative model. When constrained on this manifold, the adversarial perturbations resemble face-like features. We expect that the responses to the face-like features are effectively related for different face recognition models, which improves the transferability of the adversarial patches. This hypothesis will be confirmed in experiments.

The performance of the algorithms depends on the generative model $h$ and the latent space $\mathcal{S}$ that define the manifold. In Sec.~\ref{sec:ablation study-GenAP}, we perform ablation studies on the architectures, parameters of the generative models $h$, as well as the latent spaces $\mathcal{S}$.
First, the \textbf{capacity} of the latent space influences the white-box attack on the substitute model. The latent space of the generative model should have sufficient capacity so that the optimization can find effective adversarial examples on the white-box substitute model. Second, we observe that a generative model, which can generate features semantically related to the adversarial task at hand (\ie face features in our case), can effectively relate the responses from the substitute models and the target models and provide better \textbf{regularity}.

A straightforward algorithm to solve the optimization problem~\eqref{equ:black-box objective - latent space} is to use the Adam optimizer~\cite{kingma2014adam}.
We denote this algorithm as GenAP. Similar with TAP-TIDIM, existing techniques~\cite{xie2019improving} can be incorporated. This algorithm is depicted in Alg.~\ref{alg:G} and is denoted as GenAP-DI\footnote{GenAP-DI is Generative Adversarial Patch with Diversified Inputs}.

\begin{algorithm}[t]
\small
\caption{Transferable Adversarial Patch: GenAP-DI}
\label{alg:G}
\begin{algorithmic}[1]
\Require The adversarial objective function $\mathcal{L}_g$; a real face image $\mathbf{x}_s$ of the attacker; a  real face images $\mathbf{x}_t$ of the target identity; a binary mask matrix $\mathbf{M}$.
\Require A generative model $h$.
\Require A set of transformations $\mathcal{T}$; iterations $N$; a gradient-based optimizer, \eg, Adam~\cite{kingma2014adam}.
\Ensure
An adversarial image $\mathbf{x}^*$ by solving Eq.~\eqref{equ:black-box objective - latent space}.
\State Randomly initialize the latent vector $\mathbf{s}_0^*\sim N(0, \mathbb{I})$;
\For {$n = 0$ to $N-1$}
\State Sample a transformation $T$ from $\mathcal{T}$;
\State Blend the adversarial patch to $\mathbf{x}_s$
\begin{align}
\mathbf{x}_{n}^* &= \mathbf{x}_s \odot (1 - \mathbf{M}) + h(\mathbf{s}_n^*) \odot \mathbf{M};\nonumber
\end{align}
\State Input $T(\mathbf{x}_n^*)$ and obtain the loss $\mathcal{L}_g(T(\mathbf{x}_n^*), \mathbf{x}_t) $
\State Obtain the gradient $\nabla_{\mathbf{s}=\mathbf{s}_n^*}\mathcal{L}_{g}(T(\mathbf{x}_n^*))$;
\State Update $\mathbf{s}_{n+1}^*$ using the optimizer
\EndFor
\Return $\mathbf{x}^* = \mathbf{x}_s \odot (1 - \mathbf{M}) + h(\mathbf{s}_N^*) \odot \mathbf{M}$.
\end{algorithmic}
\end{algorithm}

\section{Experiments}

In the experiments, we demonstrate the superiority of the proposed GenAP methods in black-box attacks. Sec.~\ref{sec:setting} introduces the experimental setting. Sec.~\ref{sec:experimental results} presents the results in the digital-world attack setting. Sec.~\ref{sec:ablation study-TAP} and~\ref{sec:ablation study-GenAP} perform ablation studies on the TAP and the GenAP algorithms respectively. Sec.~\ref{sec:physical world experiment} presents the physical-world results.

\subsection{Experimental setting}\label{sec:setting}
\noindent\textbf{Datasets.}
Two face image datasets are used for evaluation: LFW~\cite{LFWTech} and CelebA-HQ~\cite{karras2017progressive}. LFW is a dataset for unconstrained face recognition. CelebA-HQ is a human face dataset of high quality. The datasets are used to test the generalization of our methods on both low quality and high quality face images, as the generative models we used are essentially trained on high quality images.

For each dataset, we select face image pairs to evaluate the adversarial algorithms on that dataset. For face verification, we select $400$ pairs in dodging attack, where each pair belongs to the same identity, and another $400$ pairs in impersonation attack, where the images from the same pair are from different identities. For face identification, we select $400$ images of $400$ different identities as the gallery set, and the corresponding $400$ images of the same identities to form the probe set. Both dodging and impersonation are performed on the probe set. This setting follows~\cite{dong2019efficient}.

\noindent\textbf{Face recognition models.}
We study three face recognition models, including FaceNet~\cite{schroff2015facenet}, CosFace~\cite{wang2018cosface} and ArcFace~\cite{deng2019arcface}, which all achieve over $99\%$ accuracies on the LFW validation set.
In testing, the feature representation for each input face image is extracted. Then, the cosine similarity between a pair of face images is calculated and compared with a threshold. We first calculate the threshold of each face recognition model by the LFW validation set. It contains $6,000$ pairs of images from same identities $(3,000)$ and different identities $(3,000)$. We choose the threshold of each model that gives the highest accuracy on this validation set. In addition, we also evaluate the performance of our method on commercial face recognition systems---Face++ and Aliyun. Given a pair of face images, a system returns a score indicating their similarity.

\begin{figure}
    \centering
    \subfigure[Eyeglass frame]
    {
    \includegraphics[width=2.0cm]{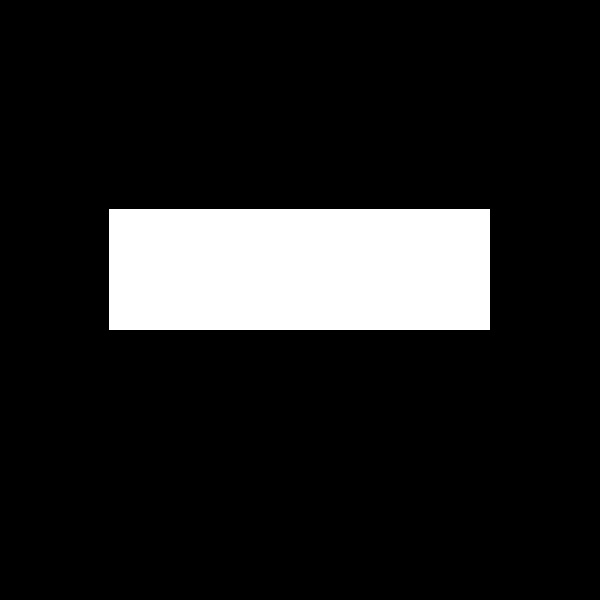}
    }\qquad
    \subfigure[Respirator]
    {
    \includegraphics[width=2.0cm]{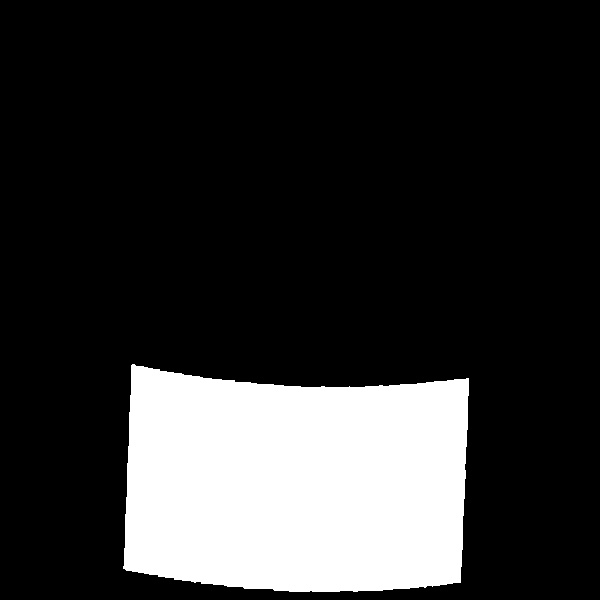}
    }
    \caption{The binary masks $\mathbf{M}$ indicating the regions of the designed patches. (a) An eyeglass frame. (b) A respirator.}
    \label{fig:demo_areas}
    \vspace{-.3cm}
\end{figure}

\noindent\textbf{Generative models.} We study three pre-trained generative models, including ProGAN~\cite{karras2017progressive}, StyleGAN~\cite{karras2019style} and StyleGAN2~\cite{karras2020analyzing}, which can generate face images of high quality. ProGAN has only one latent space. For StyleGANs, we use the $\mathbf{Z}$, $\mathbf{W}$, $\mathbf{W}^+$ and the noise latent spaces ~\cite{abdal2019image2stylegan,karras2019style}.

\noindent\textbf{Regions of patches.} We use two different regions to generate the patches, an eyeglass frame and a respirator, to show the generalization of the proposed methods to different face regions. The binary masks indicating the regions of these patches are displayed in Fig.~\ref{fig:demo_areas}.

\noindent\textbf{Evaluate Metrics.}
We use the thresholding strategy and nearest neighbor classifier for face verification and identification, respectively. To evaluate the attack performance, we report the success rate (higher is better) as the fraction of adversarial images that are not classified to the attacker himself by the model in dodging attack, and are misclassified to the desired target identity in impersonation attack.

\subsection{Experimental results}\label{sec:experimental results}
In this section, we present the experimental results of adversarial patches for black-box attack in the digital world. We generate adversarial patches using the TAP and the GenAP algorithms, respectively. In impersonation attack, we also use a vanilla baseline of pasting the corresponding face region from the target identity to the attacker (PASTE).
We then feed the generated adversarial examples to our local models and commercial APIs to test the success rates of attacks. For the TAP algorithms, we use $\epsilon=40$, which achieves the best transferability as shown by the ablation study in Sec.~\ref{sec:ablation study-TAP}. For the GenAP algorithms, we use StyleGAN2 and its $W^+$ plus the noise space as a representative, where the results of other generative models and latent spaces are left to ablation studies in Sec.~\ref{sec:ablation study-GenAP}.
We show the results on the face verification task using the eyeglass frame in Tab.~\ref{tab:dodging-verification-eyeglass} (dodging) and~\ref{tab:impersonate-verification-eyeglass} (impersonation). Results on face identification, the respirator mask and SemanticAdv~\cite{qiu2020semanticadv} are in the supplementary materials, which are qualitatively similar with the results in Tab.~\ref{tab:dodging-verification-eyeglass} and~\ref{tab:impersonate-verification-eyeglass}.

The results show that the adversarial patches achieve high success rates on the black-box models. First, TAP-TIDIM outperforms TAP-MIM. This shows that applying the existing techniques~\cite{dong2018boosting,xie2019improving,dong2019evading} originally proposed to improve the transferability of $\mathcal{L}_p$-norm constrained adversarial examples against image classification models can be helpful for improving the transferability of adversarial patches against face recognition models as well. Second, the vanilla GenAP significantly outperforms TAP algorithms in most cases (except when using FaceNet as the substitute model for impersonation attack) without bells and whistles. These results show the effectiveness of the proposed regularization method to improve the transferability of the patches. Third, the vanilla GenAP and the more sophisticated GenAP-DI performs similarly, showing that applying additional techniques~\cite{xie2019improving} do not necessarily significantly improve the performance of the GenAP algorithms.
Forth, the GenAP algorithms significantly outperform PASTE. This shows that the GenAP algorithms do not naively generating the face features of the target identity, but search the optimal adversarial face features fitting the attacker's own face features.
Fifth, our results also show the insecurity of the commercial systems (Face++ and Aliyun) against adversarial patches.

\begin{table*}[t]
    \begin{center}
    \small
    \newcommand{\tabincell}[2]{\begin{tabular}{@{}#1@{}}#2\end{tabular}}
    \tabcolsep=0.11cm
    \begin{tabular}{c|c|c|c|c|c|c|c|c|c|c|c}
    \hline
        \multirow{2}{*}{} & \multirow{2}{*}{Attack} & \multicolumn{5}{c|}{CelebA-HQ} & \multicolumn{5}{c}{LFW} \\
         \cline{3-12}
         &  & ArcFace & CosFace & FaceNet & Face++ & Aliyun &
              ArcFace & CosFace & FaceNet & Face++ & Aliyun  \\
         \hline
         ArcFace& \tabincell{c}{TAP-MIM \\ TAP-TIDIM \\ GenAP (ours) \\ GenAP-DI (ours)} &
        \tabincell{c}{$0.9875^*$\\$\mathbf{1.0000}^*$\\$0.9975^*$\\$\mathbf{1.0000}^*$\\} &
        \tabincell{c}{$0.1800$\\$0.2975$\\$\mathbf{0.6100}$\\$0.5050$\\} &
        \tabincell{c}{$0.6475$\\$0.7050$\\$\mathbf{0.9375}$\\$0.8600$\\} &
        \tabincell{c}{$0.0000$\\$0.1625$\\$\mathbf{0.7975}$\\$0.6600$\\} &
        \tabincell{c}{$0.1800$\\$0.7350$\\$\mathbf{0.9900}$\\$0.9800$\\} &
        \tabincell{c}{$0.9850^*$\\$\mathbf{1.0000}^*$\\$0.9975^*$\\$\mathbf{1.0000}^*$\\} &
        \tabincell{c}{$0.1475$\\$0.2500$\\$\mathbf{0.4850}$\\$0.4050$\\} &
        \tabincell{c}{$0.4275$\\$0.5200$\\$\mathbf{0.8725}$\\$0.7725$\\} &
        \tabincell{c}{$0.0075$\\$0.1550$\\$\mathbf{0.7450}$\\$0.6250$\\} &
        \tabincell{c}{$0.1850$\\$0.5850$\\$\mathbf{0.9850}$\\$\mathbf{0.9850}$\\}
        \\
         \hline
         CosFace& \tabincell{c}{TAP-MIM \\ TAP-TIDIM \\ GenAP \\ GenAP-DI} &
        \tabincell{c}{$0.0475$\\$0.0025$\\$\mathbf{0.5375}$\\$0.3650$\\} &
        \tabincell{c}{$0.9925^*$\\$\mathbf{1.0000}^*$\\$0.9975^*$\\$\mathbf{1.0000}^*$\\} &
        \tabincell{c}{$0.6950$\\$0.4925$\\$\mathbf{0.9550}$\\$0.9150$\\} &
        \tabincell{c}{$0.0025$\\$0.0350$\\$\mathbf{0.5675}$\\$0.3675$\\} &
        \tabincell{c}{$0.4250$\\$0.6550$\\$\mathbf{1.0000}$\\$0.9950$\\} &
        \tabincell{c}{$0.0125$\\$0.0100$\\$\mathbf{0.3650}$\\$0.2175$\\} &
        \tabincell{c}{$0.9975^*$\\$\mathbf{1.0000}^*$\\$\mathbf{1.0000}^*$\\$\mathbf{1.0000}^*$\\} &
        \tabincell{c}{$0.4950$\\$0.3550$\\$\mathbf{0.9275}$\\$0.9025$\\} &
        \tabincell{c}{$0.0075$\\$0.0425$\\$\mathbf{0.5600}$\\$0.3900$\\} &
        \tabincell{c}{$0.5350$\\$0.5650$\\$\mathbf{0.9900}$\\$0.9800$\\}
        \\
         \hline
         FaceNet& \tabincell{c}{TAP-MIM \\ TAP-TIDIM \\ GenAP \\ GenAP-DI} &
        \tabincell{c}{$0.0250$\\$0.0025$\\$\mathbf{0.3575}$\\$0.2100$\\} &
        \tabincell{c}{$0.2100$\\$0.1525$\\$\mathbf{0.3475}$\\$0.1950$\\} &
        \tabincell{c}{$0.9900^*$\\$\mathbf{0.9975}^*$\\$\mathbf{0.9975}^*$\\$\mathbf{0.9975}^*$\\} & \tabincell{c}{$0.0025$\\$0.0775$\\$\mathbf{0.5675}$\\$0.4400$\\} &
        \tabincell{c}{$0.2350$\\$0.7050$\\$\mathbf{1.0000}$\\$0.9800$\\} &
        \tabincell{c}{$0.0075$\\$0.0025$\\$\mathbf{0.1950}$\\$0.0950$\\} &
        \tabincell{c}{$0.1475$\\$0.1025$\\$\mathbf{0.2450}$\\$0.1500$\\} &
        \tabincell{c}{$0.9675^*$\\$\mathbf{1.0000}^*$\\$0.9950^*$\\$0.9925^*$\\} &
        \tabincell{c}{$0.0050$\\$0.0850$\\$\mathbf{0.5550}$\\$0.3625$\\} &
        \tabincell{c}{$0.3400$\\$0.6300$\\$\mathbf{0.9950}$\\$0.9700$\\}
        \\
         \hline
         \hline
    \end{tabular}
    \end{center}
    \caption{The success rates of black-box dodging attack on FaceNet, CosFace, ArcFace, Face++ and Aliyun in the digital world under the face verification task. The adversarial examples are generated against FaceNet, CosFace, and ArcFace by restricting the adversarial patches to an eyeglass frame region. $^*$ indicates white-box attacks.}
    \label{tab:dodging-verification-eyeglass}
\end{table*}

\begin{table*}[t]
    \begin{center}
    \small
    \newcommand{\tabincell}[2]{\begin{tabular}{@{}#1@{}}#2\end{tabular}}
    \tabcolsep=0.11cm
    \begin{tabular}{c|c|c|c|c|c|c|c|c|c|c|c}
    \hline
        \multirow{2}{*}{} & \multirow{2}{*}{Attack} & \multicolumn{5}{c|}{CelebA-HQ} & \multicolumn{5}{c}{LFW} \\
         \cline{3-12}
         &  & ArcFace & CosFace & FaceNet & Face++ & Aliyun &
              ArcFace & CosFace & FaceNet & Face++ & Aliyun \\
         \hline
         & \tabincell{c}{PASTE} &
         \tabincell{c}{0.4725} & \tabincell{c}{0.3700} & \tabincell{c}{0.3000} &
         \tabincell{c}{0.2425} & \tabincell{c}{0.0900} &
         \tabincell{c}{0.4150} & \tabincell{c}{0.3100} & \tabincell{c}{0.1825}
         & \tabincell{c}{0.1775} & \tabincell{c}{0.0250}\\
         \hline
         ArcFace & \tabincell{c}{TAP-MIM \\ TAP-TIDIM \\ TAP-TIDIMv2 \\ GenAP (ours) \\  GenAP-DI (ours)} &
        \tabincell{c}{$0.9900^*$\\$\mathbf{1.0000}^*$\\$\mathbf{1.0000}^*$\\$\mathbf{1.0000}^*$\\$\mathbf{1.0000}^*$\\} &
        \tabincell{c}{$0.3100$\\$0.3675$\\$0.4975$\\$\mathbf{0.5825}$\\$0.5300$\\} &
        \tabincell{c}{$0.2325$\\$0.2725$\\$0.3425$\\$\mathbf{0.4625}$\\$0.4100$\\} &
        \tabincell{c}{$0.1250$\\$0.1900$\\$0.2525$\\$0.3425$\\$\mathbf{0.3500}$\\} &
        \tabincell{c}{$0.0400$\\$0.0650$\\$0.0750$\\$\mathbf{0.1700}$\\$0.1450$\\} &
        \tabincell{c}{$\mathbf{1.0000}^*$\\$\mathbf{1.0000}^*$\\$\mathbf{1.0000}^*$\\$\mathbf{1.0000}^*$\\$\mathbf{1.0000}^*$\\} &
        \tabincell{c}{$0.2600$\\$0.3125$\\$0.4225$\\$\mathbf{0.5000}$\\$0.4325$\\} &
        \tabincell{c}{$0.1875$\\$0.2025$\\$0.2350$\\$\mathbf{0.4000}$\\$0.3275$\\} &  \tabincell{c}{$0.0425$\\$0.0800$\\$0.1250$\\$\mathbf{0.2125}$\\$0.1825$\\} &
        \tabincell{c}{$0.0100$\\$0.0150$\\$0.0050$\\$\mathbf{0.1000}$\\$0.0550$\\}
        \\
         \hline
         CosFace& \tabincell{c}{TAP-MIM \\ TAP-TIDIM \\ TAP-TIDIMv2 \\ GenAP \\ GenAP-DI} &
         \tabincell{c}{$0.4275$\\$0.4550$\\$0.5250$\\$\mathbf{0.6575}$\\$0.6325$\\} &
        \tabincell{c}{$0.9900^*$\\$\mathbf{1.0000}^*$\\$\mathbf{1.0000}^*$\\$\mathbf{1.0000}^*$\\$\mathbf{1.0000}^*$\\} &
        \tabincell{c}{$0.3125$\\$0.3725$\\$0.4175$\\$0.5250$\\$\mathbf{0.5325}$\\} &
        \tabincell{c}{$0.1425$\\$0.2000$\\$0.2650$\\$\mathbf{0.3500}$\\$0.3275$\\} &
        \tabincell{c}{$0.0450$\\$0.0750$\\$0.0950$\\$\mathbf{0.2000}$\\$0.1900$\\} &
        \tabincell{c}{$0.3250$\\$0.2725$\\$0.3625$\\$\mathbf{0.5350}$\\$0.4975$\\} &
        \tabincell{c}{$0.9850^*$\\$\mathbf{1.0000}^*$\\$\mathbf{1.0000}^*$\\$0.9975$\\$\mathbf{1.0000}^*$\\} &
        \tabincell{c}{$0.2525$\\$0.2700$\\$0.3225$\\$0.4600$\\$\mathbf{0.4650}$\\} &
        \tabincell{c}{$0.0550$\\$0.0750$\\$0.1325$\\$\mathbf{0.2100}$\\$0.2000$\\} &
        \tabincell{c}{$0.0200$\\$0.0150$\\$0.0100$\\$0.0700$\\$\mathbf{0.1000}$\\}
        \\
         \hline
         FaceNet& \tabincell{c}{TAP-MIM \\ TAP-TIDIM \\ TAP-TIDIMv2 \\ GenAP \\ GenAP-DI} &
         \tabincell{c}{$0.2400$\\$0.1800$\\$0.3000$\\$0.2750$\\$0.2175$\\} &
        \tabincell{c}{$0.2025$\\$0.2200$\\$0.3300$\\$0.2450$\\$0.2025$\\} &
        \tabincell{c}{$0.8300^*$\\$\mathbf{0.9775}^*$\\$\mathbf{0.9775}^*$\\$0.9025^*$\\$0.9650^*$\\} &
        \tabincell{c}{$0.1150$\\$0.1175$\\$0.1725$\\$0.1250$\\$0.1150$\\} &
        \tabincell{c}{$0.0450$\\$0.0450$\\$0.0500$\\$0.0600$\\$0.0500$\\} &
        \tabincell{c}{$0.1425$\\$0.0925$\\$0.1650$\\$0.2450$\\$0.1675$\\} &
        \tabincell{c}{$0.1850$\\$0.1925$\\$0.2450$\\$0.2425$\\$0.1850$\\} &
        \tabincell{c}{$0.8375^*$\\$0.9800^*$\\$0.9825^*$\\$0.9200^*$\\$\mathbf{0.9850}^*$\\} &
        \tabincell{c}{$0.0300$\\$0.0300$\\$0.0650$\\$0.0900$\\$0.0550$\\} &
        \tabincell{c}{$0.0100$\\$0.0050$\\$0.0150$\\$\mathbf{0.0350}$\\$\mathbf{0.0350}$\\}
         \\
         \hline
         \hline
    \end{tabular}
    \end{center}
    \caption{The success rates of black-box impersonation attack on FaceNet, CosFace, ArcFace, Face++ and Aliyun in the digital world under the face verification task. The adversarial examples are generated against FaceNet, CosFace, and ArcFace by restricting the adversarial patches to an eyeglass frame region. $^*$ indicates white-box attacks.}
    \label{tab:impersonate-verification-eyeglass}
    \vspace{-.3cm}
\end{table*}

\subsection{Ablation study on TAP-TIDIM}\label{sec:ablation study-TAP}
This section presents the ablation studies on the TAP algorithms to support the discussions in Sec.~\ref{sec:pixel space optimization}. These ablation studies show that TAP-TIDIM has trouble escaping local optima of unsatisfactory transferability, though many regularization methods have been used~\cite{xie2019improving,dong2019evading}. This motivates us to develop new regularization in this paper.

\subsubsection{Sensitivity to initialization}
The initialization step is the only difference between TAP-TIDIM and TAP-TIDIMv2 when solving problem~\eqref{equ:black-box objective}. But TAP-TIDIMv2 shows significantly higher success rates in black-box impersonation attack, as shown in Tab.~\ref{tab:impersonate-verification-eyeglass}. While the solution of TAP-TIDIMv2 is within the search space of TAP-TIDIM\footnote{Strictly speaking, the solution of TAP-TIDIMv2 is within the search space of TAP-TIDIM ($\epsilon=255$) and the results in Tab.~\ref{tab:impersonate-verification-eyeglass} are for TAP-TIDIM ($\epsilon=40$). Nevertheless, the TAP-TIDIM ($\epsilon=40$) outperforms TAP-TIDIM ($\epsilon=255$) as shown in Fig.~\ref{fig:epislon} and our conclusion holds.}, TAP-TIDIM cannot find it and is trapped into local optima with significantly worse transferability.

\subsubsection{Sensitivity to $\epsilon$}\label{sec:epsilon}
\begin{figure}[t]
\begin{center}
\includegraphics[width=0.7\linewidth]{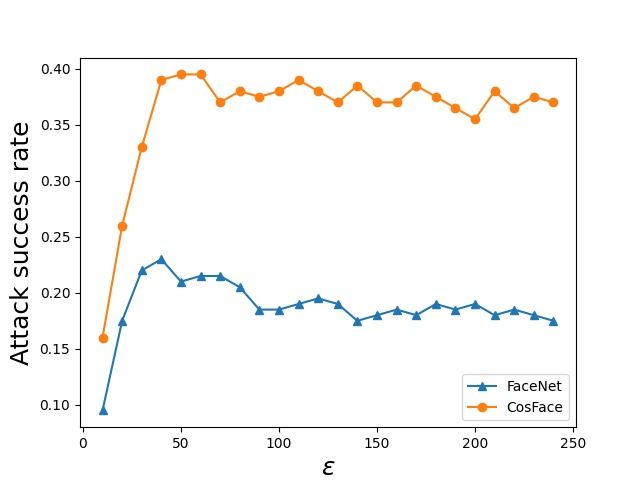}
\vspace{-.3cm}
\end{center}
   \caption{The success rates of TAP-TIDIM on the black-box models first rise and then fall when the maximal perturbation magnitude $\epsilon$ increases. This indicates that the adversarial patches are overfitting the substitute model. The results are black-box impersonation attack on FaceNet and CosFace under the face verification task. The adversarial examples are generated against ArcFace by restricting the adversarial patches to an eyeglass frame region. $200$ image pairs from the LFW dataset are used.}
\label{fig:epislon}
\vspace{-.3cm}
\end{figure}

The hyperparameter $\epsilon$ in TAP-TIDIM can control the upper bound for the perturbation magnitudes of the adversarial patches, which is an indicator of the size of the search space. The larger the $\epsilon$, the larger the search space.
Fig.~\ref{fig:epislon} shows that, as the upper bound $\epsilon$ increases, the success rates on the black-box models first rise and then fall. When $\epsilon$ is small, the transferability benefits from the larger search space by finding more effective adversarial examples against the substitute model. But when $\epsilon$ is large, the adversarial patches overfit the substitute model and are trapped into poor local optima. 
The transferability of TAP-TIDIM reaches the optimality around $\epsilon=40$, which is used in Sec.~\ref{sec:experimental results}.

\subsection{Ablation study on GenAP}\label{sec:ablation study-GenAP}
In this section, we present the ablation studies on the GenAP algorithms, which show how the regularity and the capacity of the manifold influence the transferability of the generative adversarial patches. This section use the GenAP algorithm  and the eyeglass frame mask for experiments.

\subsubsection{Parameters of the generative models}\label{sec:ablation-parameter}
We use the StyleGAN2 with different parameters, including the parameters that are randomly initialized (Rand), trained on the LSUN car dataset (CAR) and trained on the FFHQ face image dataset (FFHQ). We use $\mathbf{W}^+$ plus the noise space~\cite{abdal2019image2stylegan,karras2019style} as the latent space. Tab.~\ref{tab:ablation}(a) shows that using the randomly initialized model cannot find effective adversarial examples even in the white-box case. While the StyleGAN2 trained on CAR is effective at white-box attack, their transferability to black-box models is poor.
The transferability of GenAP is better than the TAP algorithms (c.f., Tab.~\ref{tab:impersonate-verification-eyeglass}) only when the generative model is trained on human face dataset. This phenomenon indicates that using face-like features as perturbations is important for bridging the gap between the substitute and the target face recognition models to improve transferability in the GenAP methods.

\begin{table*}[ht!]
    \begin{center}
    \small
    \newcommand{\tabincell}[2]{\begin{tabular}{@{}#1@{}}#2\end{tabular}}
    \begin{tabular}{c|c|c|c|c|c|c|c}
    \hline
         \multirow{2}{*}{} & \multirow{2}{*}{} & \multicolumn{3}{c|}{CelebA-HQ} & \multicolumn{3}{c}{LFW} \\\hline
         \cline{3-8}
          &  & ArcFace & CosFace & FaceNet & ArcFace & CosFace & FaceNet \\
         \hline
         \multicolumn{8}{c}{(a) Parameters}\\
         \hline
         ArcFace& \tabincell{c}{RAND \\ CAR \\ FFHQ} &
         \tabincell{c}{$0.2625^*$\\$0.9850^*$\\{$\mathbf{1.0000^*}$}} &
         \tabincell{c}{0.0100\\0.0950\\\textbf{0.5828}} &
         \tabincell{c}{0.0100\\0.1075\\\textbf{0.4625}} &
         \tabincell{c}{$0.2575^*$\\$0.9825^*$\\{$\mathbf{1.0000^*}$}} &
         \tabincell{c}{0.0050\\0.0625\\\textbf{0.5000}} &
         \tabincell{c}{0.0200\\0.0800\\\textbf{0.4000}} \\
         \hline
         \multicolumn{8}{c}{(b) Architectures}\\ \hline
         ArcFace& \tabincell{c}{ProGAN \\ StyleGAN \\ StyleGAN2} &
         \tabincell{c}{$0.6750^*$\\$\mathbf{1.0000^*}$\\{$\mathbf{1.0000^*}$}} &
         \tabincell{c}{0.2375\\0.5500\\\textbf{0.5825}} &
         \tabincell{c}{0.2125\\0.4250\\\textbf{0.4625}} &
         \tabincell{c}{$0.6475^*$\\$\mathbf{1.0000^*}$\\{$\mathbf{1.0000^*}$}} &
         \tabincell{c}{0.1650\\0.4750\\\textbf{0.5000}} &
         \tabincell{c}{0.1450\\0.3475\\\textbf{0.4000}} \\
         \hline
         \multicolumn{8}{c}{(c) Latent spaces}\\
         \hline
         ArcFace&
         \tabincell{c}{$\mathbf{Z}$ \\$\mathbf{W}$ \\ $\mathbf{W}^+$ \\ Noise \\ $\mathbf{W}^+$ + Noise} &
         \tabincell{c}{$0.4175^*$\\$0.9575^*$\\$\mathbf{1.0000}^*$\\$0.2075^*$\\{$\mathbf{1.0000^*}$}} &
        \tabincell{c}{$0.1125$\\$0.4775$\\$0.5750$\\$0.0500$\\\textbf{0.5825}\\} &
        \tabincell{c}{$0.0800$\\$0.3825$\\\textbf{0.4625}\\$0.0450$\\\textbf{0.4625}\\} &
        \tabincell{c}{$0.4050^*$\\$0.9425^*$\\$\mathbf{1.0000}^*$\\$0.1250^*$\\{$\mathbf{1.0000^*}$}\\} &
        \tabincell{c}{$0.1200$\\$0.4300$\\$0.4925$\\$0.0425$\\\textbf{0.5000}\\} &
        \tabincell{c}{$0.0900$\\$0.3900$\\$0.3825$\\$0.0250$\\\textbf{0.4000}\\}  \\
        \hline
    \end{tabular}
    \end{center}
    \caption{The success rates of black-box impersonation attack when the architectures, the parameter and the latent space are changed in the proposed GenAP algorithm. The adversarial examples are generated against ArcFace by restricting the adversarial patches to an eyeglass frame region, and are tested on FaceNet, CosFace and ArcFace in the digital world under the face verification task. $^*$ indicates white-box attacks. The ablation studies are on (a) the parameters (RAND, CAR and FFHQ) of the StyleGAN2, (b) the architectures of the generative model (ProGAN, StyleGAN, StyleGAN2) and (c) the latent space ($\mathbf{Z},\mathbf{W},\mathbf{W^+}$ and noise) used by StyleGAN2 trained on FFHQ.}
    \label{tab:ablation}
\end{table*}

\subsubsection{Architectures of the generative models}\label{sec:ablation-architecture}
We use three different generative models, including ProGAN~\cite{karras2017progressive}, StyleGAN~\cite{karras2019style} and StyleGAN2~\cite{karras2020analyzing}. These generative models differ in their network architectures and can generate human face images with higher and higher quality. For the StyleGANs, we use the $\mathbf{W}^+$ latent plus the noise space~\cite{abdal2019image2stylegan,karras2019style}. The results are shown in Tab.~\ref{tab:ablation}(b). The performance of GenAP depends on the architectures of the generative models. Even though ProGAN is trained on human face images, it is difficult to find effective adversarial examples in its latent space, even in the white-box case.
Both StyleGANs achieve high success rates against the black-box models. These phenomena indicate that the style-based decoder in StyleGANs might be important for the GenAP algorithms to find effective adversarial examples.

\subsubsection{Latent spaces of the generative models}
We use different latent spaces for the StyleGAN2, including the $\mathbf{Z}$, the $\mathbf{W}$, the $\mathbf{W}^+$ and the noise spaces~\cite{abdal2019image2stylegan,karras2019style}. The $\mathbf{W}^+$ is more flexible than the $\mathbf{Z}$, the $\mathbf{W}$ and the noise spaces with much more degrees of freedom. Tab.~\ref{tab:ablation}(c) shows that, the performance on the $\mathbf{W}$ and the $\mathbf{W}^+$ spaces is substantially higher than that on the $\mathbf{Z}$ and the noise spaces. The optimizations in the $\mathbf{Z}$ and the noise spaces cannot find effective adversarial patches even in the white-box case.

\subsection{Physical-world experiment}\label{sec:physical world experiment}
In this section, we verify that the adversarial patches generated by the proposed GenAP algorithms are physically realizable, and their superiority is retained after printing and photographing. Specifically, we select a volunteer as the attacker and 3 target identities (one male and two females) from the CelebA-HQ dataset. For each target identity, we generate an eyeglass frame for the attacker to impersonate that identity.
After the attacker wears the adversarial eyeglass frame, we take a video of him from the front and randomly select 100 video frames. The video frames are used for face verification. We evaluate the transferability of the patches using the cosine similarities. The higher the similarity, the better the transferability. Results in Fig.\ref{tab:physical} show that the patches generated by the proposed GenAP-DI retain high transferability even after printing and photographing.

\subsection{Extra experiments}
In the supplementary material, we also compare the proposed GenAP methods with an existing method of using face-like features as adversarial perturbations, SemanticAdv~\cite{qiu2020semanticadv}, and extend the GenAP methods to other tasks, \eg, image classification. First, we explain why SemanticAdv is sub-optimal in the patch setting. Second, we show the generalizability of the proposed GenAP methods to other recognition tasks.

\section{Conclusion}
In this paper, we evaluate the robustness of face recognition models against adversarial patches in the query-free black-box setting. Firstly, we extend existing techniques from the $\mathcal{L}_p$-constrained $(p>0)$ setting to the patch setting, yielding TAP algorithms to generate transferable adversarial patches. However, several experimental phenomena indicate that it is hard for the TAP algorithms to escape from local optima with unsatisfactory transferability. Therefore, we propose to regularize the adversarial patches on the manifold learnt by generative models pre-trained on human face images. The perturbations in the proposed GenAP algorithms resemble face-like features, which is important for reducing the gap between the substitute and the target face recognition models. Experiments confirm the superiority of the proposed methods.

\begin{table}
    \begin{center}
    \small
    \newcommand{\tabincell}[2]{\begin{tabular}{@{}#1@{}}#2\end{tabular}}
    \begin{tabular}{c|c|c|c}
    \hline
          & target 1 & target 2 & target 3 \\
         \hline
         \multicolumn{4}{c}{CosFace} \\
         \hline
         \tabincell{c}{TAP-TIDIMv2 \\ GenAP-DI (ours)} &
         \tabincell{c}{$16.8(\pm 1.6)$ \\ $\mathbf{27.2}(\pm 2.2)$} &
         \tabincell{c}{$18.2(\pm 1.3)$ \\ $\mathbf{21.4}(\pm 2.3)$ } &
         \tabincell{c}{$4.1(\pm 2.0)$ \\ $\mathbf{12.0}(\pm 2.4)$} \\
         \hline
         \multicolumn{4}{c}{FaceNet} \\
         \hline
         \tabincell{c}{TAP-TIDIMv2 \\ GenAP-DI (ours)} &
         \tabincell{c}{$\mathbf{23.3}(\pm 2.4)$\\ $22.8(\pm 2.6)$} &
         \tabincell{c}{$-3.9(\pm 3.0$ \\ $\mathbf{24.6}(\pm 2.0)$} &
         \tabincell{c}{$32.2(\pm 2.1)$ \\ $\mathbf{33.4}(\pm 1.8)$ } \\
         \hline
    \end{tabular}
    \end{center}
    \caption{The cosine similarties between the attacker wearing the adversarial eyeglass frame and three different target identities in the physical-world. The target identities are randomly drawn from CelebA-HQ. The adversarial eyeglass frame is crafted by the TAP-TIDIMv2 and the proposed GenAP-DI algorithms on ArcFace, and is tested on CosFace and FaceNet.}
    \vspace{-.3cm}
    \label{tab:physical}
\end{table}


{\small
\bibliographystyle{ieee_fullname}
\bibliography{egbib}
}

\clearpage
\appendix

\section{Implementation details}

\subsection{Hyperparameters}
\noindent\textbf{TAP algorithms:} We set the number of iterations $N=400$, the learning rate $\alpha=1$ and the decay factor $\mu=1$. We use $\epsilon=255$ for dodging and $\epsilon=40$ for impersonation.

\noindent\textbf{GenAP algorithms:} We set the number of iterations $N=100$, and the learning rate of Adam optimizer $\alpha=0.01$.

\subsection{Models}
\noindent\textbf{Face recognition models:} All face recognition models are accessible from the Internet, including FaceNet\footnote{\url{https://github.com/timesler/facenet-pytorch}}, CosFace\footnote{\url{https://github.com/MuggleWang/CosFace_pytorch}}, ArcFace\footnote{\url{https://github.com/TreB1eN/InsightFace_Pytorch}}, Face++\footnote{\url{https://www.faceplusplus.com/face-comparing/}} and Aliyun\footnote{\url{https://vision.aliyun.com/facebody}}.

\noindent\textbf{Generative models:} All generative models are accessible from the Internet, including ProGAN\footnote{\url{https://github.com/tkarras/progressive_growing_of_gans/tree/master}}, StyleGAN\footnote{\url{https://github.com/NVlabs/stylegan}}, StyleGAN2\footnote{\url{https://github.com/NVlabs/stylegan2}}.

\section{Additional experiments on adversarial eyeglass frame}
In the main text, we present the attack success rates on using adversarial eyeglass frames to perform dodging and impersonation attacks on the face verification task. In this section, we present the success rates on dodging and impersonation attacks on the face identification task in Tab.~\ref{tab:dodging-identification-eyeglass} and~\ref{tab:impersonate-identification-eyeglass}. The conclusions on these results are consistent.

\begin{table*}[t]
    \begin{center}
    \footnotesize
    \newcommand{\tabincell}[2]{\begin{tabular}{@{}#1@{}}#2\end{tabular}}

    \begin{tabular}{c|c|c|c|c|c|c|c}
    \hline
        \multirow{2}{*}{} & \multirow{2}{*}{Attack} & \multicolumn{3}{c|}{CelebA-HQ} & \multicolumn{3}{c}{LFW} \\
         \cline{3-8}
         &  & ArcFace & CosFace & FaceNet & ArcFace & CosFace & FaceNet \\
         \hline
         ArcFace& \tabincell{c}{TAP-MIM \\ TAP-TIDIM \\ GenAP \\ GenAP-DI} &
        \tabincell{c}{$0.9875^*$\\$\mathbf{1.0000}^*$\\$0.9950^*$\\$\mathbf{1.0000}^*$\\} &
        \tabincell{c}{$0.3000$\\$0.3750$\\$\mathbf{0.7100}$\\$0.5975$\\} &
        \tabincell{c}{$0.7550$\\$0.8050$\\$\mathbf{0.9650}$\\$0.9200$\\} &
        \tabincell{c}{$0.9875^*$\\$\mathbf{1.0000}^*$\\$\mathbf{1.0000}^*$\\$\mathbf{1.0000}^*$\\} &
        \tabincell{c}{$0.2150$\\$0.3250$\\$\mathbf{0.5650}$\\$0.4750$\\} &
        \tabincell{c}{$0.5950$\\$0.6625$\\$\mathbf{0.9250}$\\$0.8500$\\}
        \\
         \hline
         CosFace& \tabincell{c}{TAP-MIM \\ TAP-TIDIM \\ GenAP \\ GenAP-DI} &
        \tabincell{c}{$0.1600$\\$0.0425$\\$\mathbf{0.6700}$\\$0.5350$\\} &
        \tabincell{c}{$0.9950^*$\\$\mathbf{1.0000}^*$\\$\mathbf{1.0000}^*$\\$\mathbf{1.0000}^*$\\} &
        \tabincell{c}{$0.8175$\\$0.6700$\\$\mathbf{0.9675}$\\$0.9400$\\} &
        \tabincell{c}{$0.0525$\\$0.0275$\\$\mathbf{0.5700}$\\$0.3700$\\} &
        \tabincell{c}{$0.9975^*$\\$\mathbf{1.0000}^*$\\$\mathbf{1.0000}^*$\\$\mathbf{1.0000}^*$\\} &
        \tabincell{c}{$0.7075$\\$0.5425$\\$\mathbf{0.9850}$\\$0.9550$\\}
         \\
         \hline
         FaceNet& \tabincell{c}{TAP-MIM \\ TAP-TIDIM \\ GenAP \\ GenAP-DI} &
        \tabincell{c}{$0.1075$\\$0.0350$\\$\mathbf{0.4825}$\\$0.3425$\\} &
        \tabincell{c}{$0.2750$\\$0.1900$\\$\mathbf{0.4000}$\\$0.2650$\\} &
        \tabincell{c}{$0.9950^*$\\$\mathbf{1.0000}^*$\\$0.9975^*$\\$\mathbf{1.0000}^*$\\} &
        \tabincell{c}{$0.0425$\\$0.0125$\\$\mathbf{0.3375}$\\$0.1875$\\} &
        \tabincell{c}{$0.2100$\\$0.1475$\\$\mathbf{0.3250}$\\$0.1675$\\} &
        \tabincell{c}{$0.9900^*$\\$\mathbf{1.0000}^*$\\$0.9975^*$\\$0.9950^*$\\}
        \\
         \hline
         \hline
    \end{tabular}
    \end{center}
    \caption{The success rates of black-box dodging attack on FaceNet, CosFace, ArcFace in the digital world under the face identification task. The adversarial examples are generated against FaceNet, CosFace, and ArcFace by restricting the adversarial patches to a eyeglass frame region. $^*$ indicates white-box attacks.}
    \label{tab:dodging-identification-eyeglass}
\end{table*}

\begin{table*}[t]
    \begin{center}
    \small
    \newcommand{\tabincell}[2]{\begin{tabular}{@{}#1@{}}#2\end{tabular}}

    \begin{tabular}{c|c|c|c|c|c|c|c}
    \hline
        \multirow{2}{*}{} & \multirow{2}{*}{Attack} & \multicolumn{3}{c|}{CelebA-HQ} & \multicolumn{3}{c}{LFW} \\
         \cline{3-8}
         &  & ArcFace & CosFace & FaceNet & ArcFace & CosFace & FaceNet\\
         \hline
         ArcFace& \tabincell{c}{TAP-MIM \\ TAP-TIDIM \\ TAP-TIDIMv2 \\ GenAP \\ GenAP-DI} &
        \tabincell{c}{$0.6450^*$\\$0.9050^*$\\$0.9375^*$\\$0.8625^*$\\$\mathbf{0.9625}^*$\\} &
        \tabincell{c}{$0.1000$\\$0.1200$\\$0.2575$\\$\mathbf{0.3425}$\\$0.3225$\\} &
        \tabincell{c}{$0.0700$\\$0.1000$\\$0.1900$\\$\mathbf{0.2750}$\\$0.2425$\\} &
        \tabincell{c}{$0.6850^*$\\$0.9275^*$\\$\mathbf{0.9425}^*$\\$0.8675^*$\\$0.9400^*$\\} &
        \tabincell{c}{$0.0925$\\$0.1175$\\$0.2075$\\$\mathbf{0.2425}$\\$0.2325$\\} &
        \tabincell{c}{$0.0650$\\$0.0625$\\$0.0950$\\$\mathbf{0.2150}$\\$0.1550$\\}
        \\
         \hline
         CosFace& \tabincell{c}{TAP-MIM \\ TAP-TIDIM \\ TAP-TIDIMv2 \\ GenAP \\ GenAP-DI} &
        \tabincell{c}{$0.1150$\\$0.1200$\\$0.1700$\\$\mathbf{0.2975}$\\$0.2375$\\} &
        \tabincell{c}{$0.7425^*$\\$0.9625^*$\\$0.9875^*$\\$0.9425^*$\\$\mathbf{0.9950}^*$\\} &
        \tabincell{c}{$0.1525$\\$0.1650$\\$0.2275$\\$\mathbf{0.3625}$\\$0.3550$\\} &
        \tabincell{c}{$0.1175$\\$0.1100$\\$0.1650$\\$\mathbf{0.2750}$\\$0.2500$\\} &
        \tabincell{c}{$0.7250^*$\\$0.9750^*$\\$0.9900^*$\\$0.9350^*$\\$\mathbf{0.9925}^*$\\} &
        \tabincell{c}{$0.0825$\\$0.1075$\\$0.1500$\\$\mathbf{0.2875}$\\$0.2675$\\}
         \\
         \hline
         FaceNet& \tabincell{c}{TAP-MIM \\ TAP-TIDIM \\ TAP-TIDIMv2 \\ GenAP \\ GenAP-DI} &
        \tabincell{c}{$0.0325$\\$0.0325$\\$0.0775$\\$\mathbf{0.0900}$\\$0.0650$\\} &
        \tabincell{c}{$0.0625$\\$0.0750$\\$\mathbf{0.1500}$\\$0.1225$\\$0.1050$\\} &
        \tabincell{c}{$0.5600*$\\$0.8975^*$\\$\mathbf{0.9125}^*$\\$0.7900^*$\\$0.9025^*$\\} &
        \tabincell{c}{$0.0375$\\$0.0275$\\$0.0550$\\$\mathbf{0.1200}$\\$0.0825$\\} &
        \tabincell{c}{$0.0600$\\$0.0700$\\$0.0800$\\$\mathbf{0.1100}$\\$0.0800$\\} &
        \tabincell{c}{$0.4900^*$\\$0.9075^*$\\$\mathbf{0.9225}^*$\\$0.7450^*$\\$\mathbf{0.9225}^*$\\}
        \\
         \hline
         \hline
    \end{tabular}
    \end{center}
    \caption{The success rates of black-box impersonation attack on FaceNet, CosFace, ArcFace in the digital world under the face identification task. The adversarial examples are generated against FaceNet, CosFace, and ArcFace by restricting the adversarial patches to a eyeglass frame region. $^*$ indicates white-box attacks.}
    \label{tab:impersonate-identification-eyeglass}
\end{table*}

\begin{figure*}[htp]
\centering
\begin{tabular}{ccc}
    \includegraphics[width=0.18\linewidth]{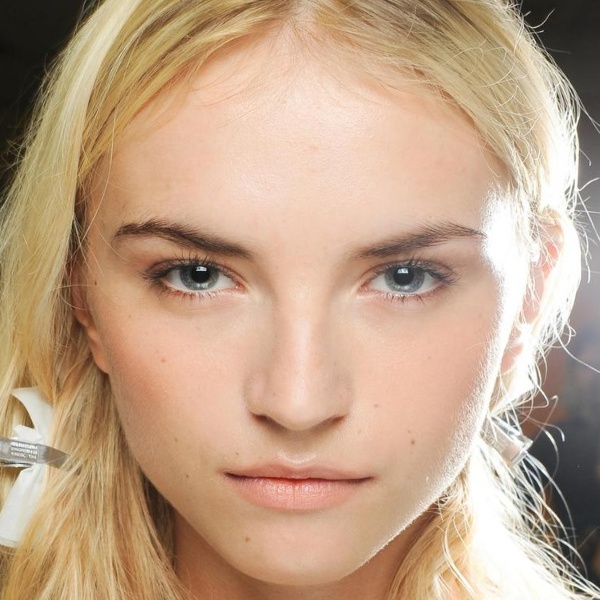} &
    \includegraphics[width=0.18\linewidth]{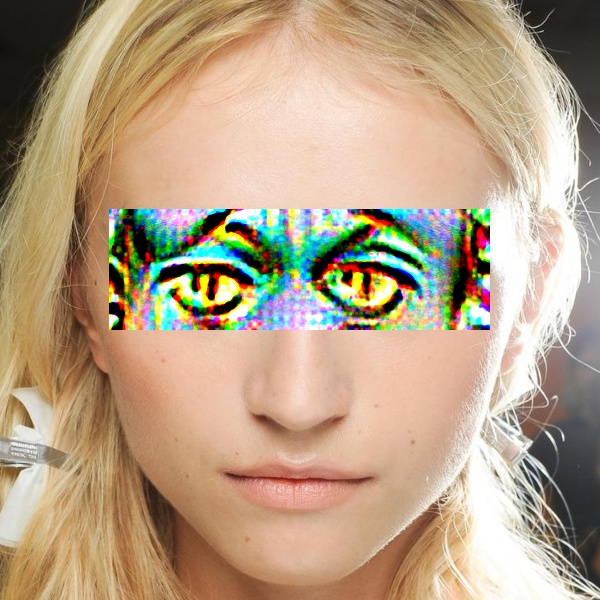} &
    \includegraphics[width=0.18\linewidth]{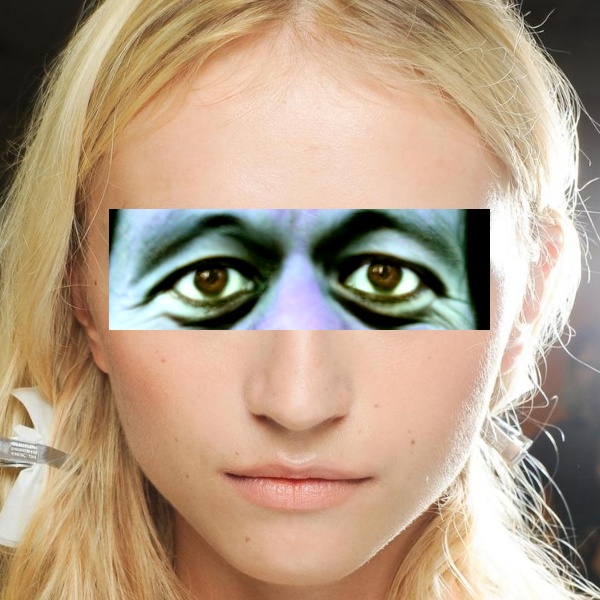}
    \\
    \includegraphics[width=0.18\linewidth]{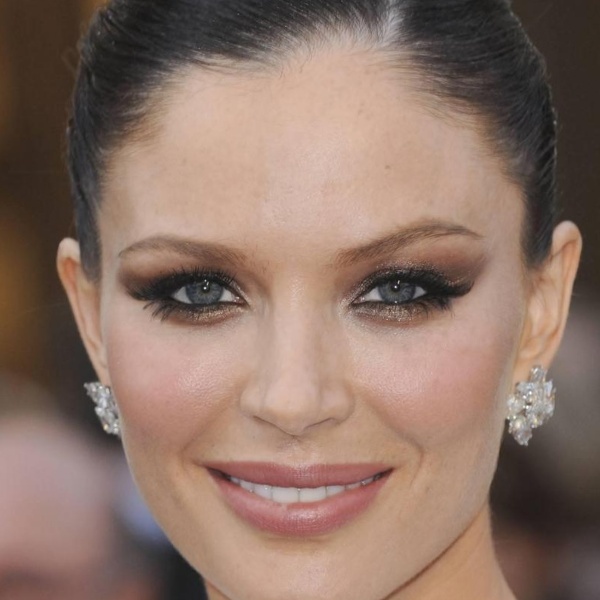} &
    \includegraphics[width=0.18\linewidth]{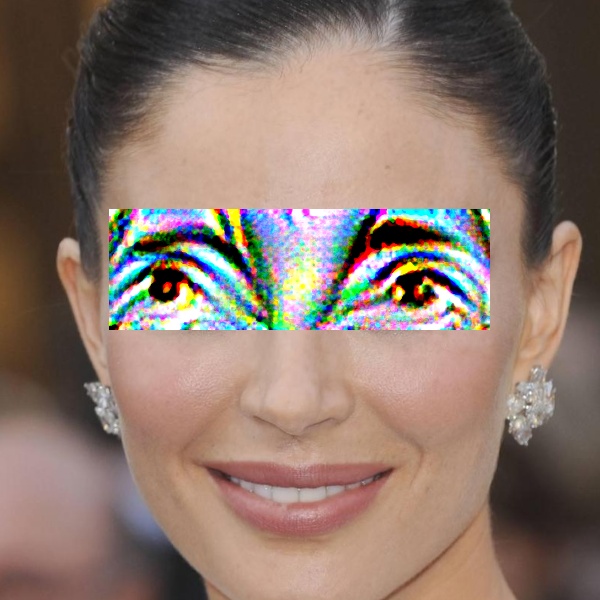} &
    \includegraphics[width=0.18\linewidth]{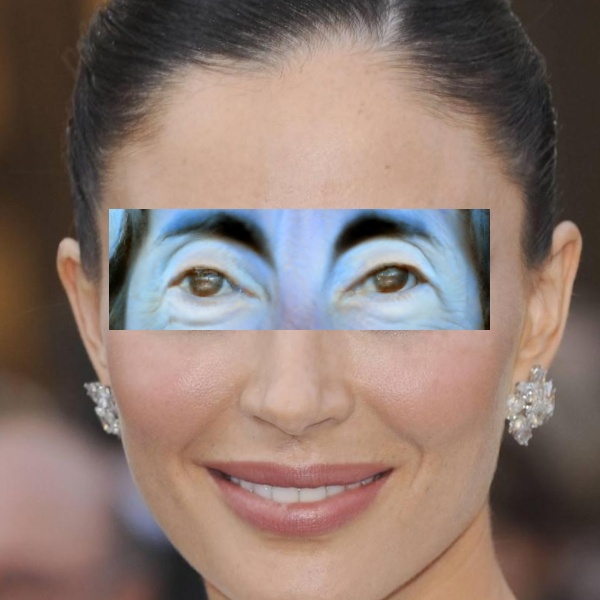}
    \\
    \includegraphics[width=0.18\linewidth]{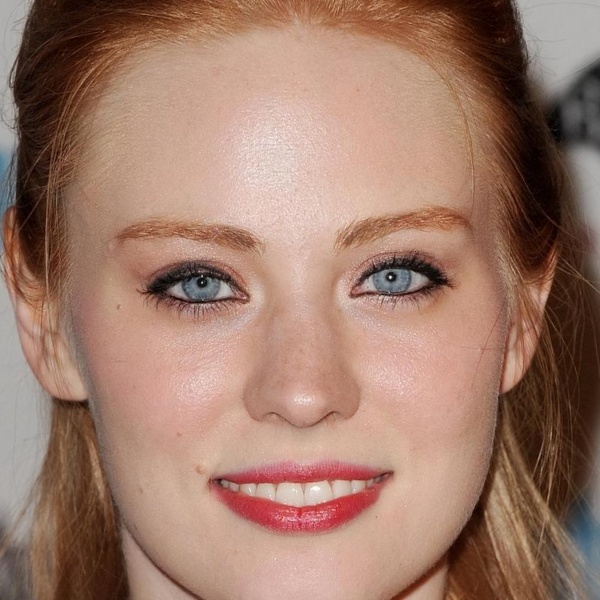} &
    \includegraphics[width=0.18\linewidth]{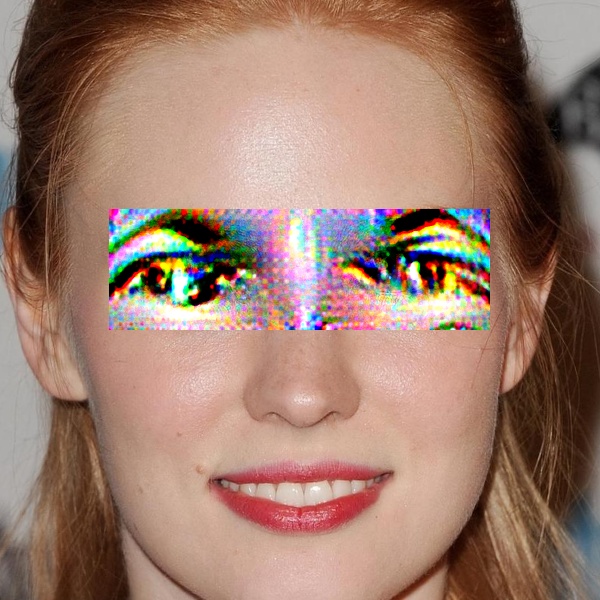} &
    \includegraphics[width=0.18\linewidth]{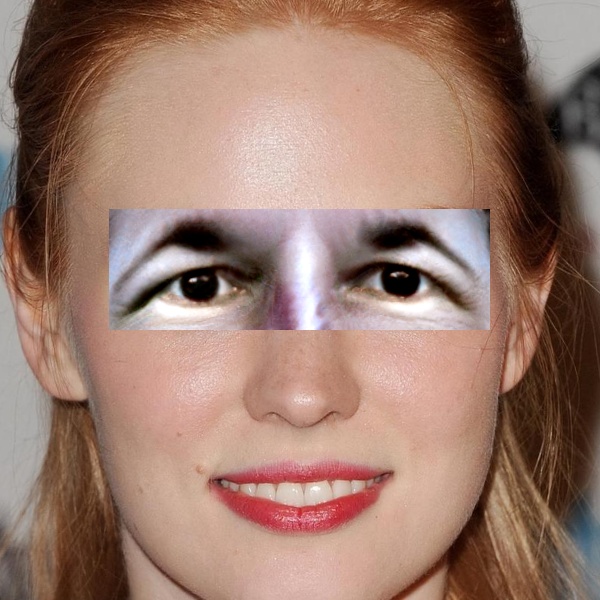}
    \\
    \includegraphics[width=0.18\linewidth]{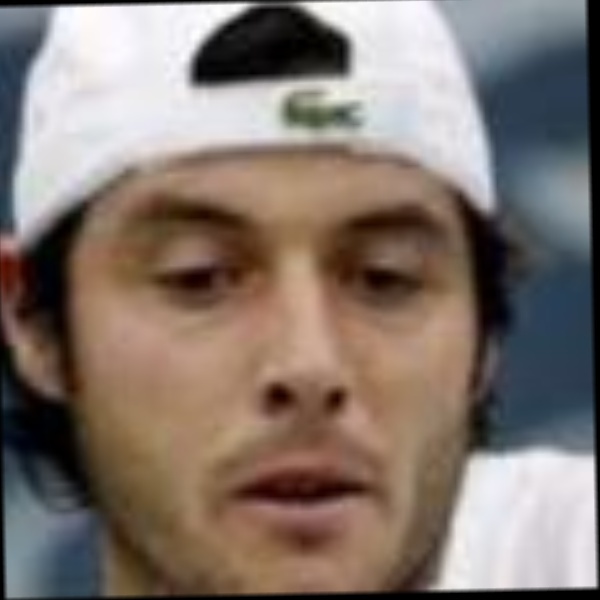} &
    \includegraphics[width=0.18\linewidth]{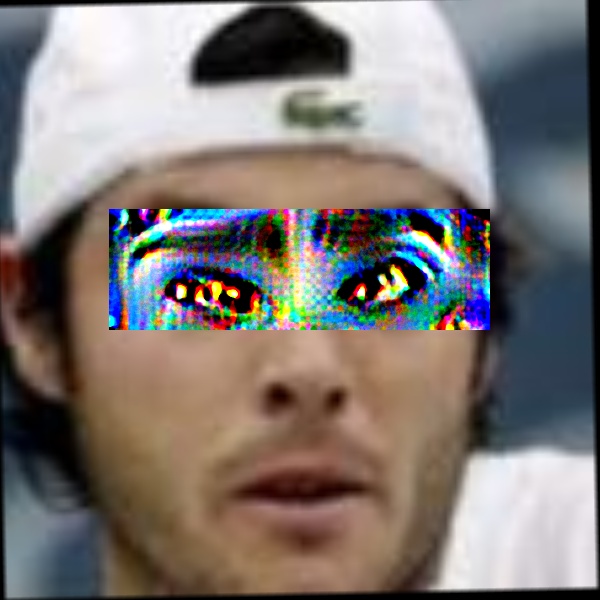} &
    \includegraphics[width=0.18\linewidth]{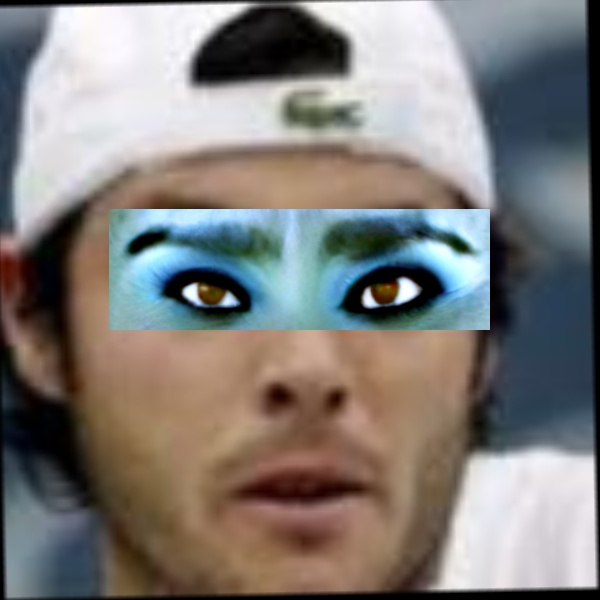}
    \\
    \includegraphics[width=0.18\linewidth]{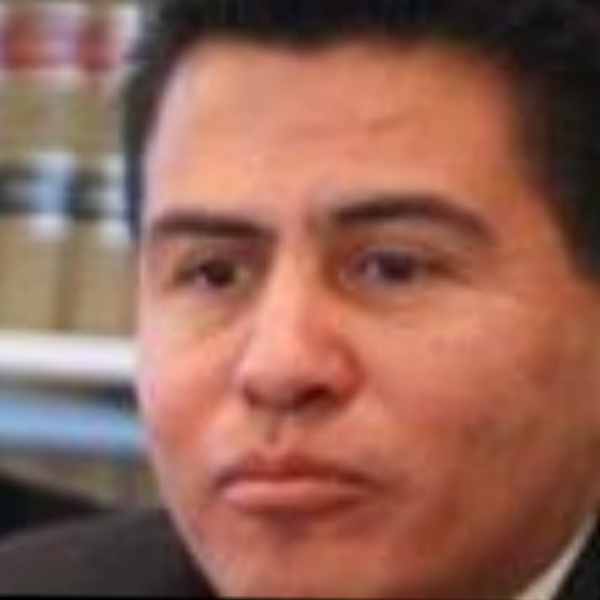} &
    \includegraphics[width=0.18\linewidth]{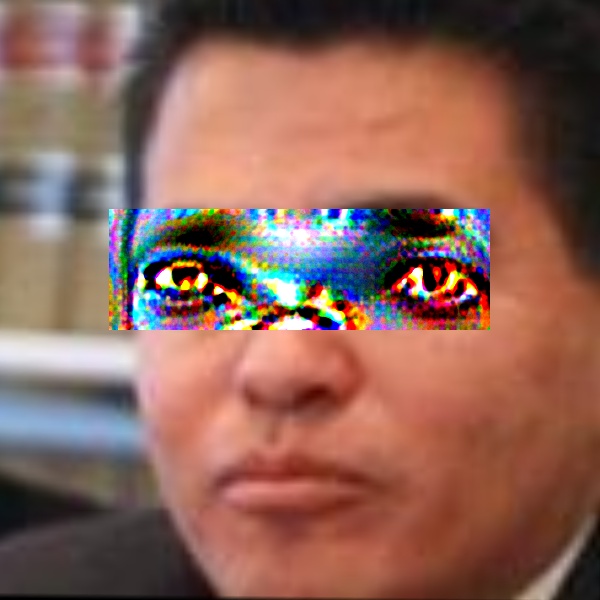} &
    \includegraphics[width=0.18\linewidth]{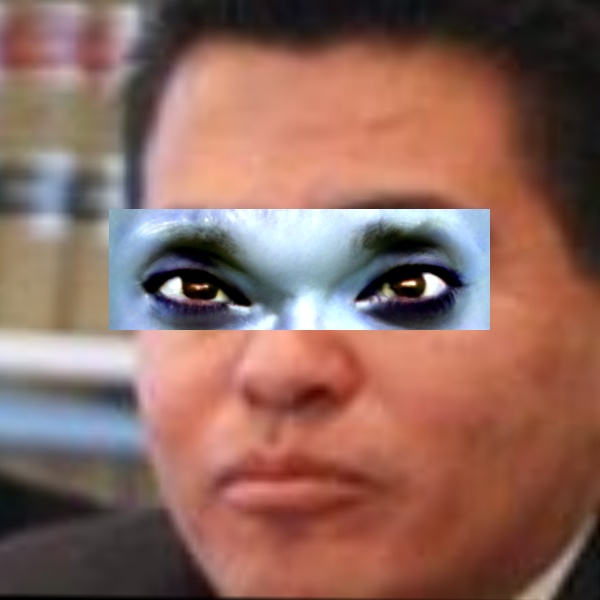}
    \\
    \includegraphics[width=0.18\linewidth]{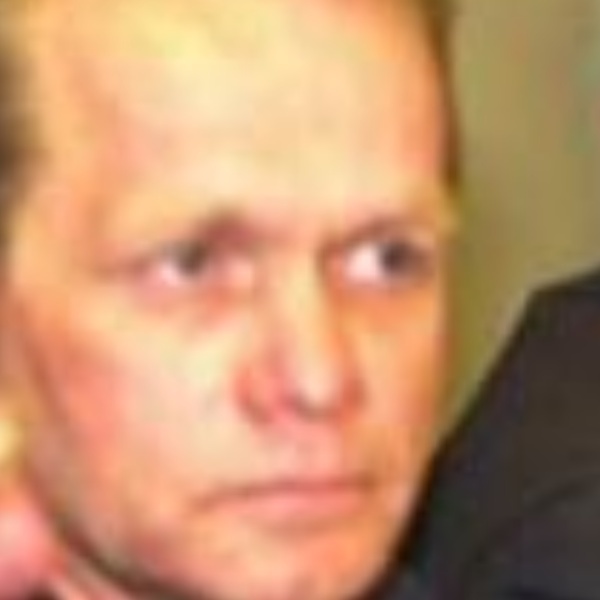} &
    \includegraphics[width=0.18\linewidth]{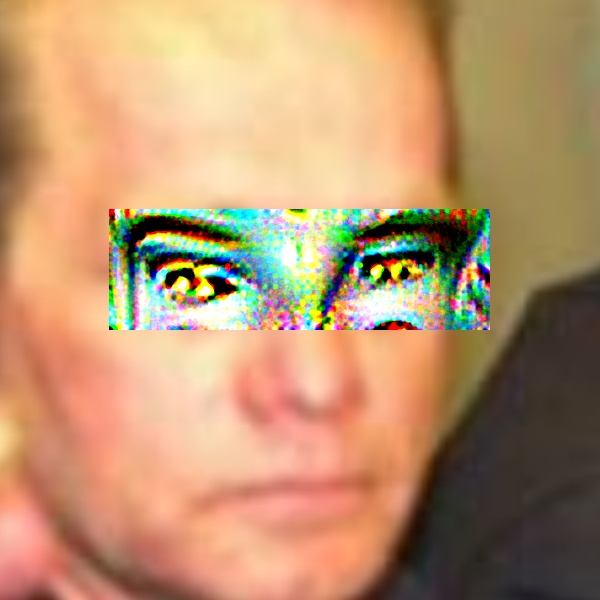} &
    \includegraphics[width=0.18\linewidth]{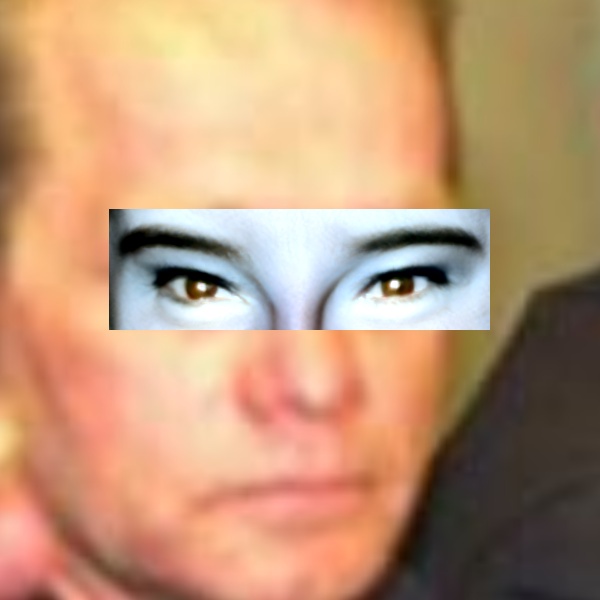}
    \\
    Attacker & TAP-TIDIM & GenAP-DI \\
\end{tabular}
\caption{Visualization of adversarial eyeglass frames generated by the TAP-TIDIM and the GenAP-DI methods for dodging attack. The first three rows are the demonstrations on CelebA-HQ dataset and the others are from LFW dataset. And the three columns denotes the pictures of attackers and attackers with the adversarial eyeglass frames by generated by TAP-TIDIM and GenAP-DI methods separately. In TAP-TIDIM, we use $\epsilon=255$.
}
\label{fig:dodging-eyeglass} \vspace{-0.5em}
\end{figure*}

\begin{figure*}[htp]
\centering
\begin{tabular}{cccc}
    \includegraphics[width=0.18\linewidth]{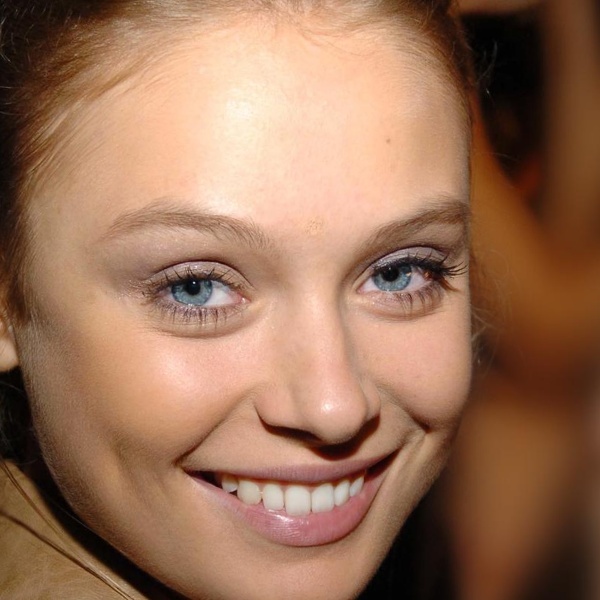} &
    \includegraphics[width=0.18\linewidth]{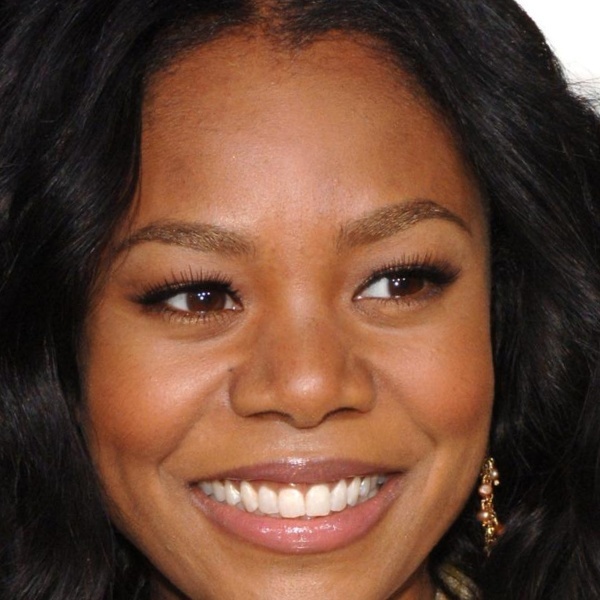} &
    \includegraphics[width=0.18\linewidth]{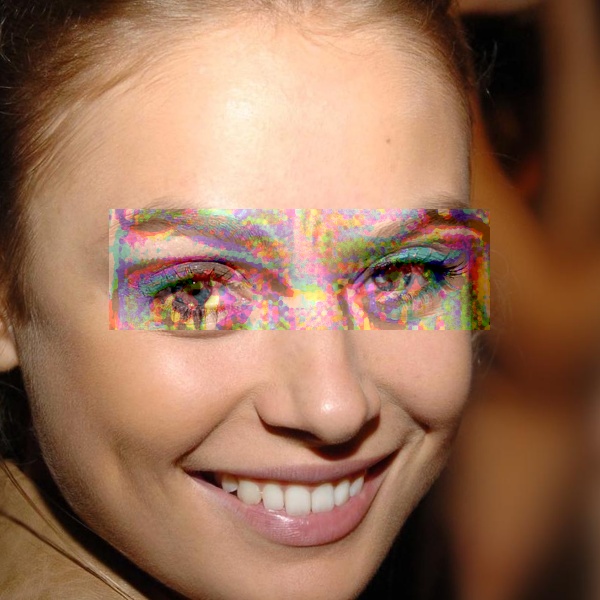} &
    \includegraphics[width=0.18\linewidth]{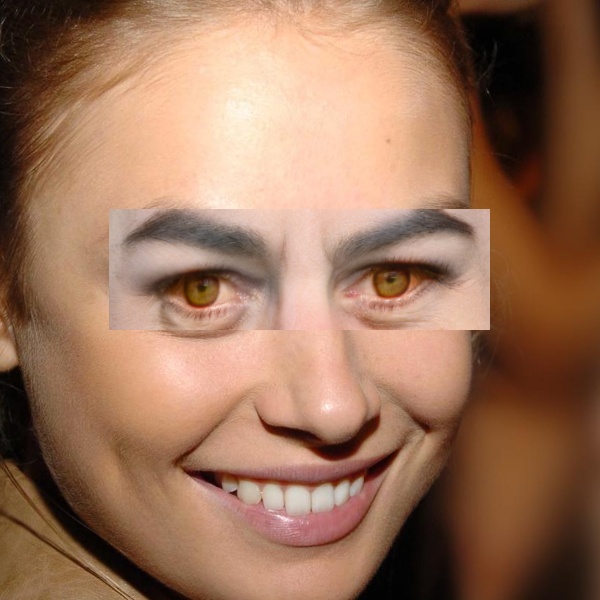}
    \\
    \includegraphics[width=0.18\linewidth]{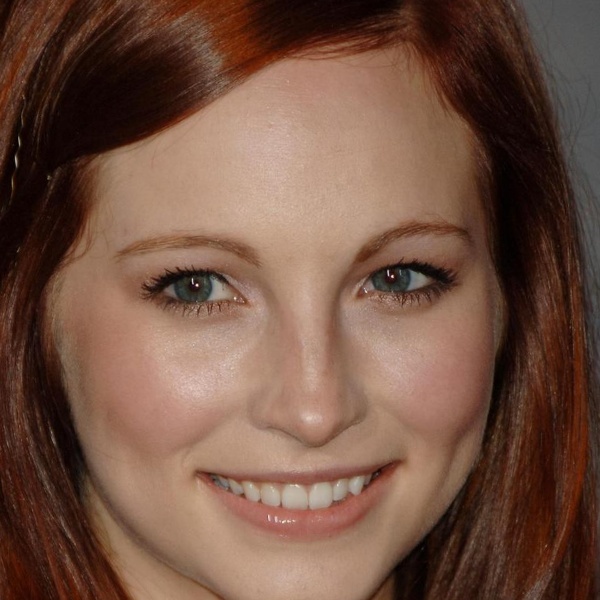} &
    \includegraphics[width=0.18\linewidth]{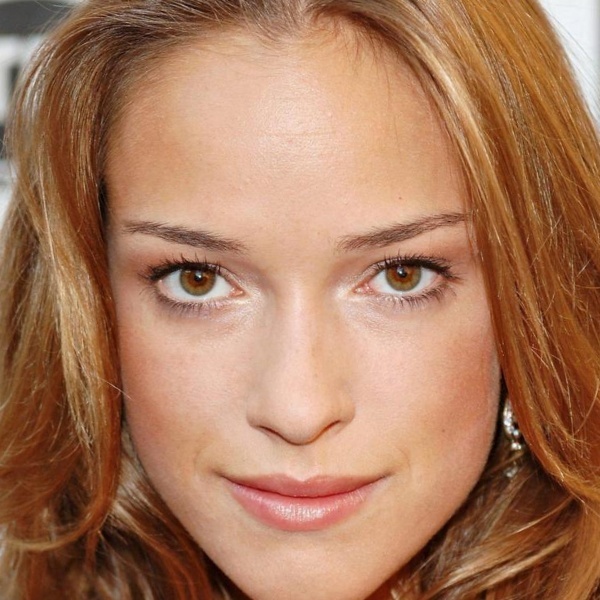} &
    \includegraphics[width=0.18\linewidth]{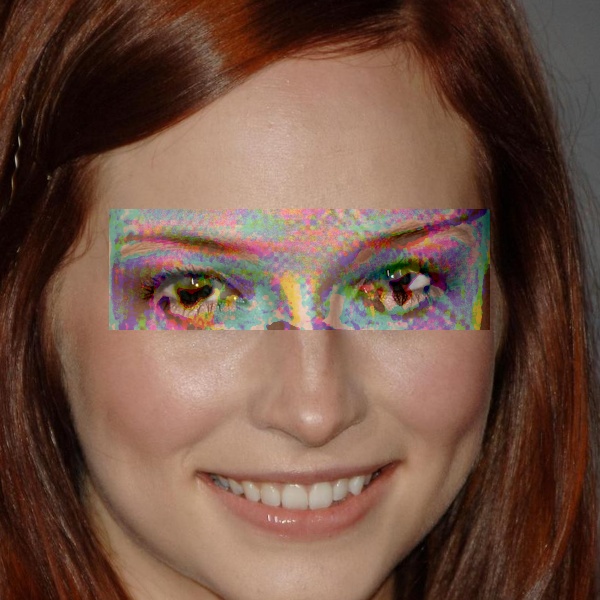} &
    \includegraphics[width=0.18\linewidth]{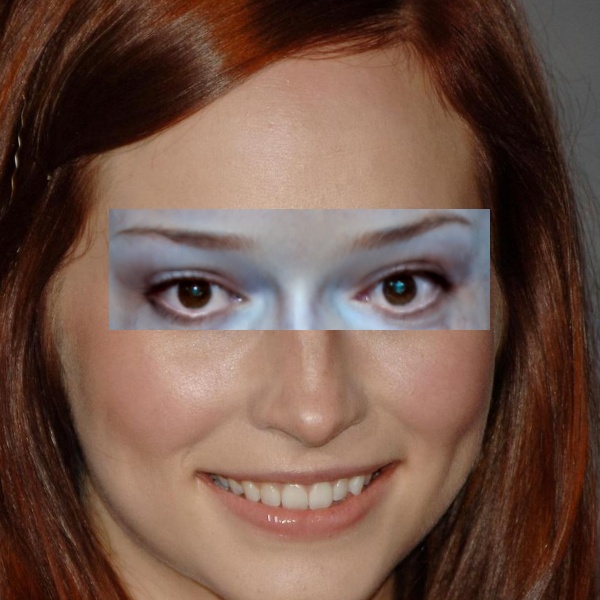}
    \\
    \includegraphics[width=0.18\linewidth]{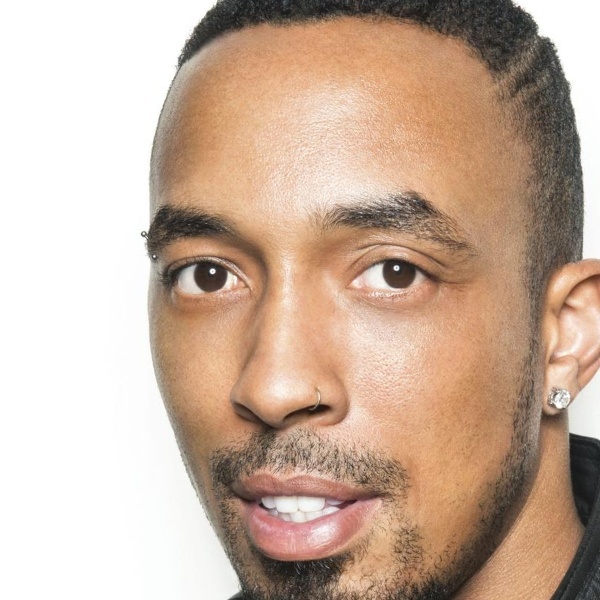} &
    \includegraphics[width=0.18\linewidth]{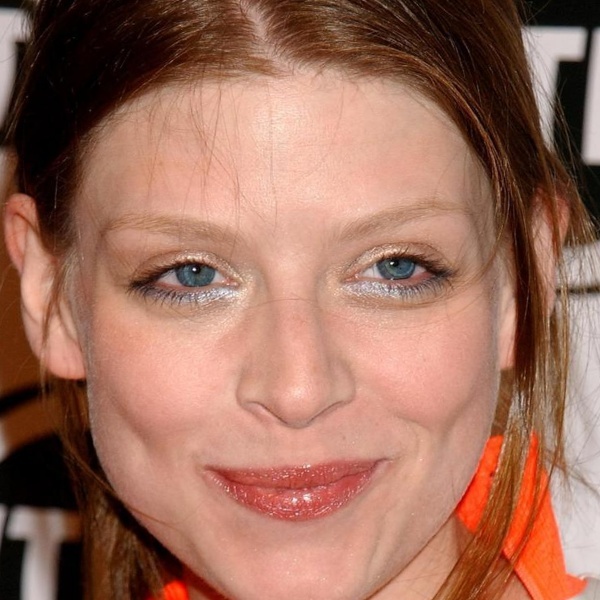} &
    \includegraphics[width=0.18\linewidth]{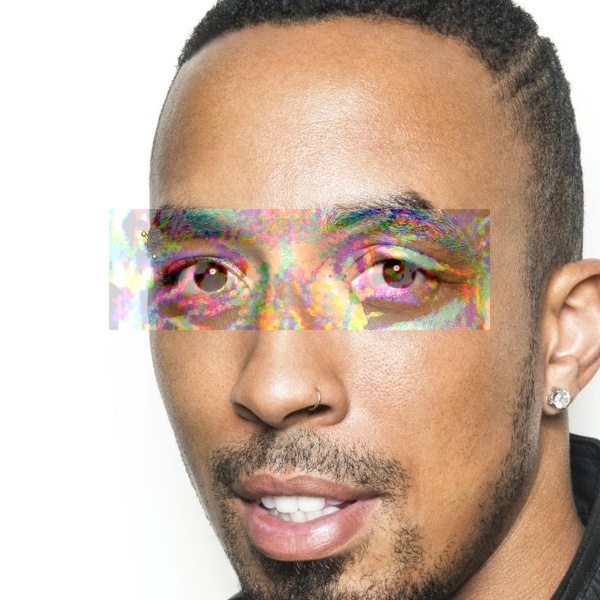} &
    \includegraphics[width=0.18\linewidth]{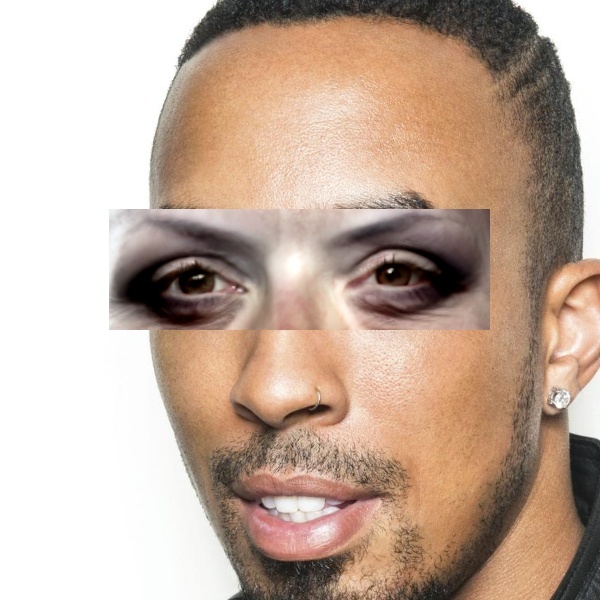}
    \\
    \includegraphics[width=0.18\linewidth]{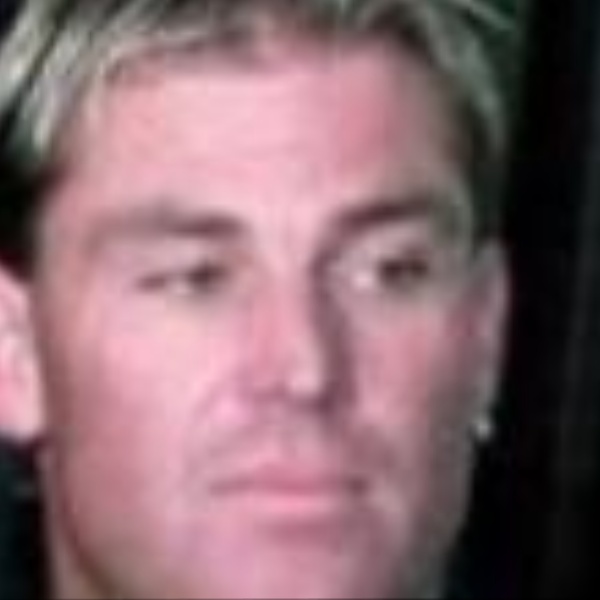} &
    \includegraphics[width=0.18\linewidth]{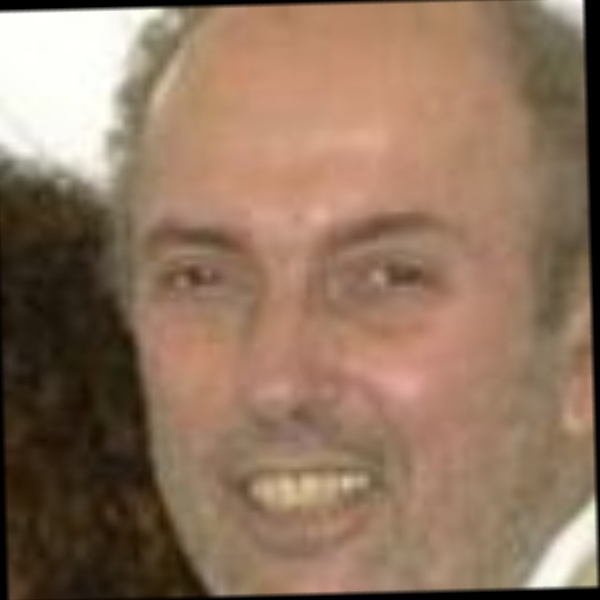} &
    \includegraphics[width=0.18\linewidth]{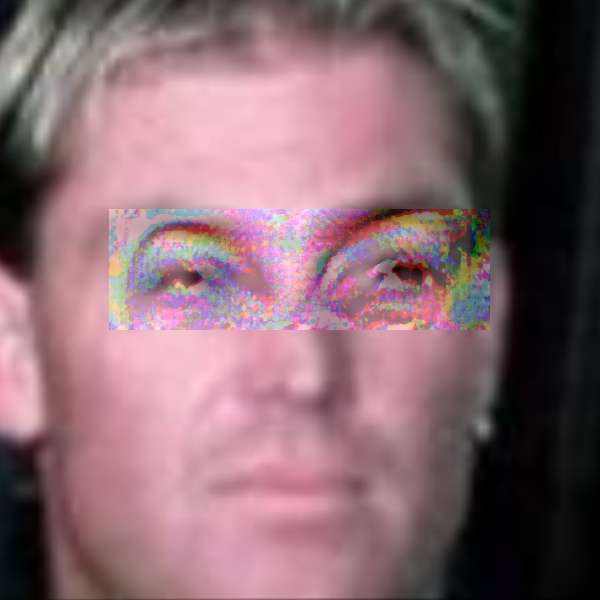} &
    \includegraphics[width=0.18\linewidth]{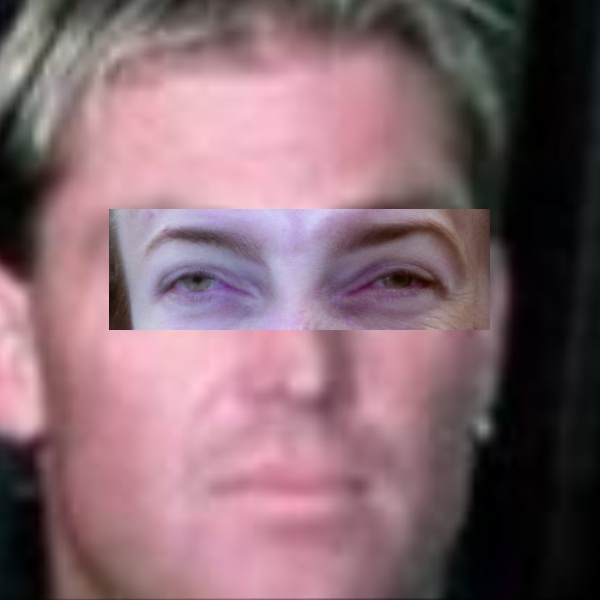}
    \\
    \includegraphics[width=0.18\linewidth]{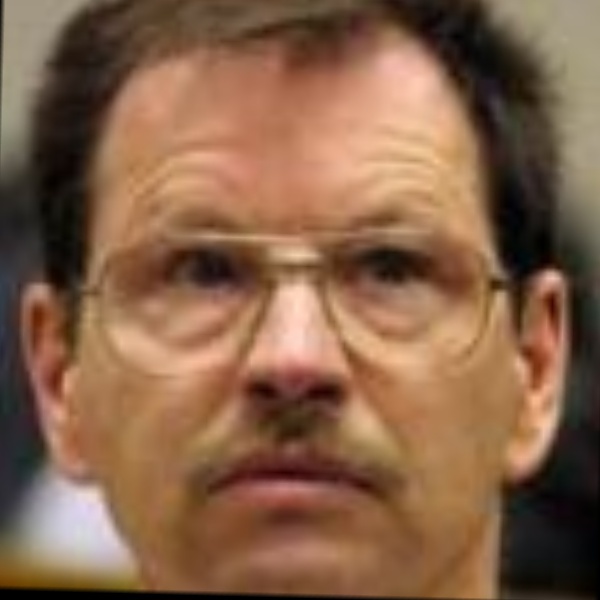} &
    \includegraphics[width=0.18\linewidth]{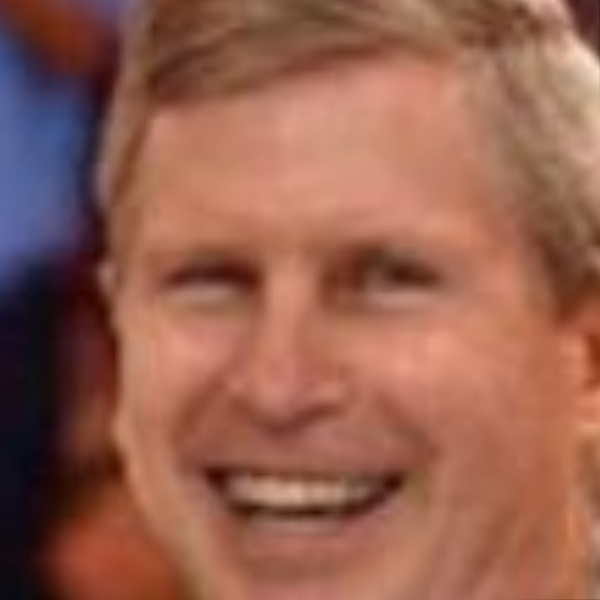} &
    \includegraphics[width=0.18\linewidth]{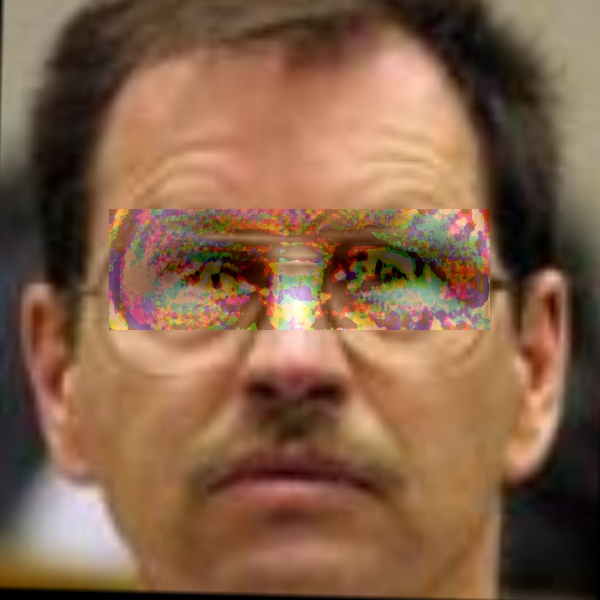} &
    \includegraphics[width=0.18\linewidth]{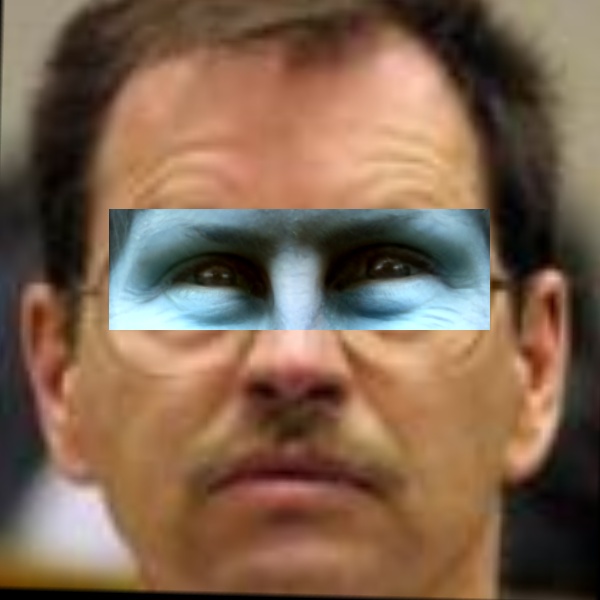}
    \\
    \includegraphics[width=0.18\linewidth]{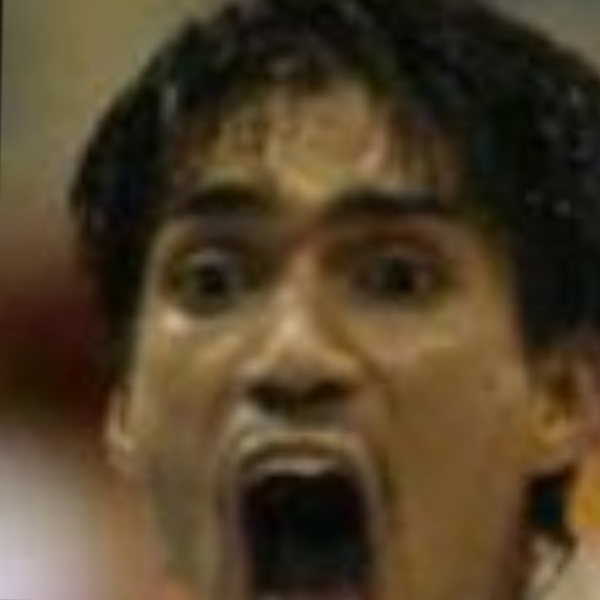} &
    \includegraphics[width=0.18\linewidth]{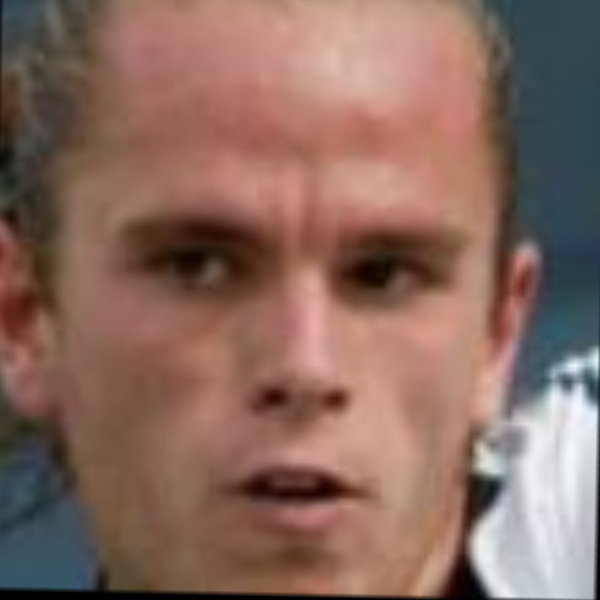} &
    \includegraphics[width=0.18\linewidth]{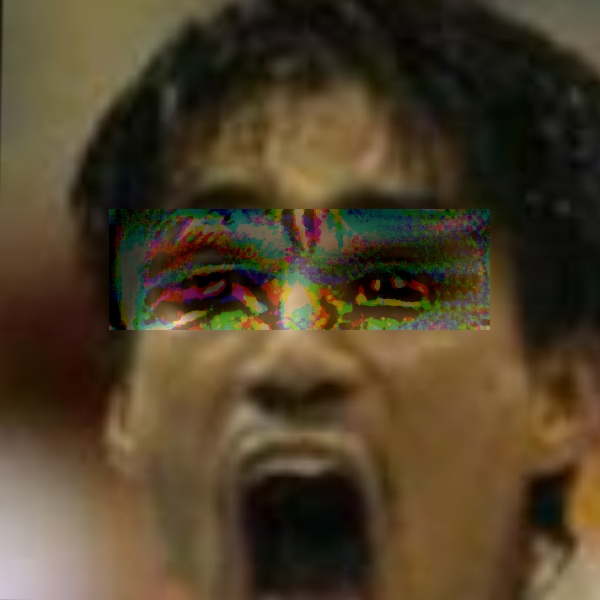} &
    \includegraphics[width=0.18\linewidth]{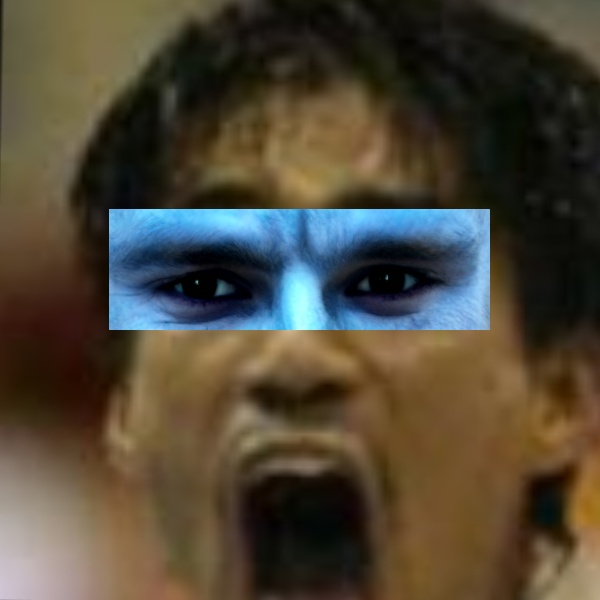}
    \\
    Attacker & Target identity & TAP-TIDIM & GenAP-DI \\
\end{tabular}
\caption{Visualization of adversarial eyeglass frames generated by the TAP-TIDIM and the GenAP-DI methods for impersonation attack. The first three rows are the demonstrations on CelebA-HQ dataset and the others are from LFW dataset. The first two columns are the photos of attackers and their target identities, and the following two columns show the attackers with the adversarial eyeglass frames generated by the proposed TAP-TIDIM and GenAP-DI methods. In TAP-TIDIM, we use $\epsilon=40$.}
\label{fig:impersonation-eyeglass} \vspace{-0.5em}
\end{figure*}

Moreover, we visualize the adversarial examples generated by the TAP-TIDIM and the GenAP-DI methods for dodging attack (see Fig.~\ref{fig:dodging-eyeglass}) and impersonation attack (see Fig.~\ref{fig:impersonation-eyeglass}), respectively. The proposed GAP methods generates face-like features as perturbations.

\section{Experiments on adversarial respirators}
\begin{figure*}[htp]
\centering
\begin{tabular}{cccc}
    \includegraphics[width=0.18\linewidth]{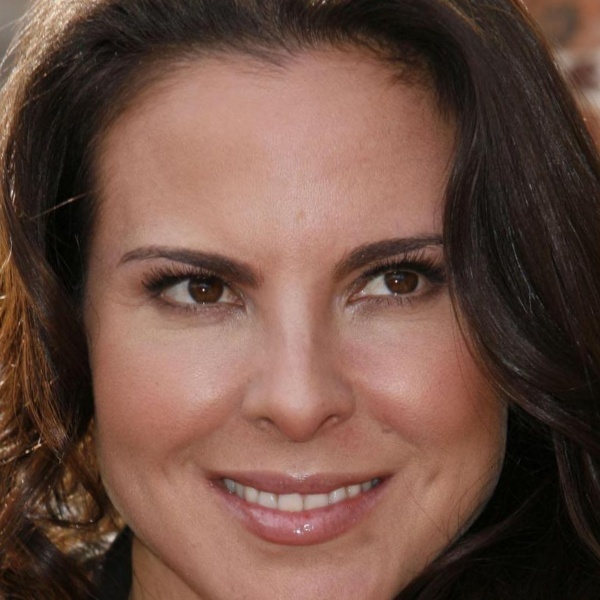} &
    \includegraphics[width=0.18\linewidth]{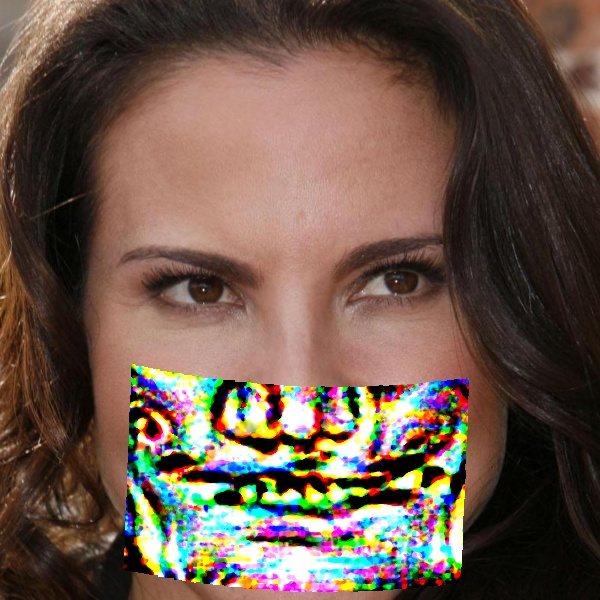} &
    \includegraphics[width=0.18\linewidth]{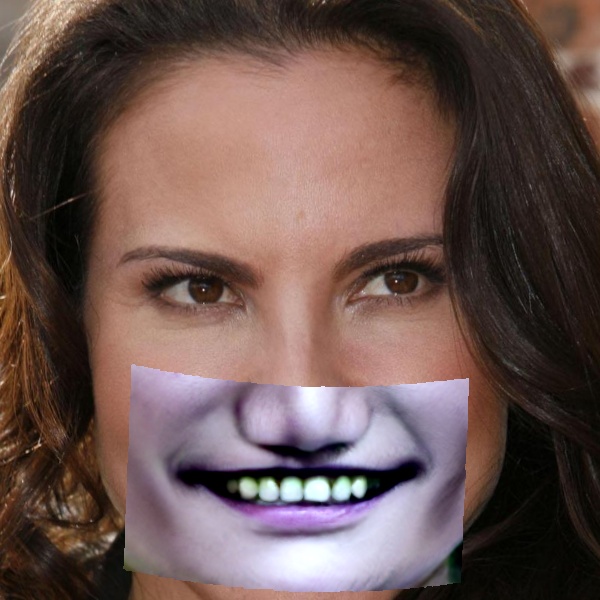}
    \\
    \includegraphics[width=0.18\linewidth]{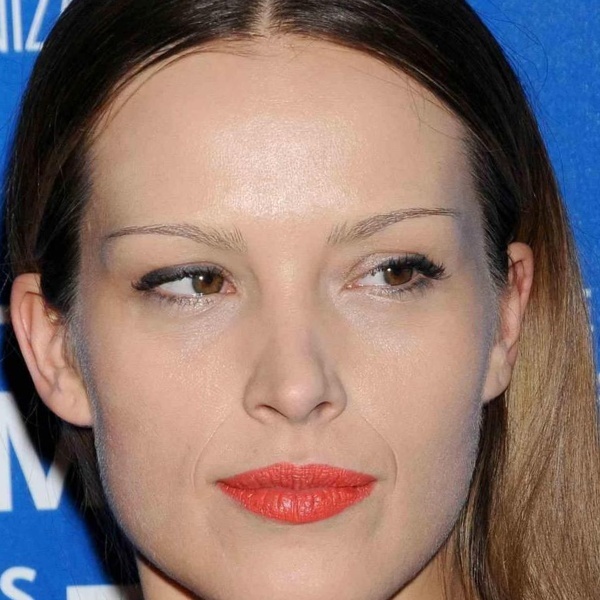} &
    \includegraphics[width=0.18\linewidth]{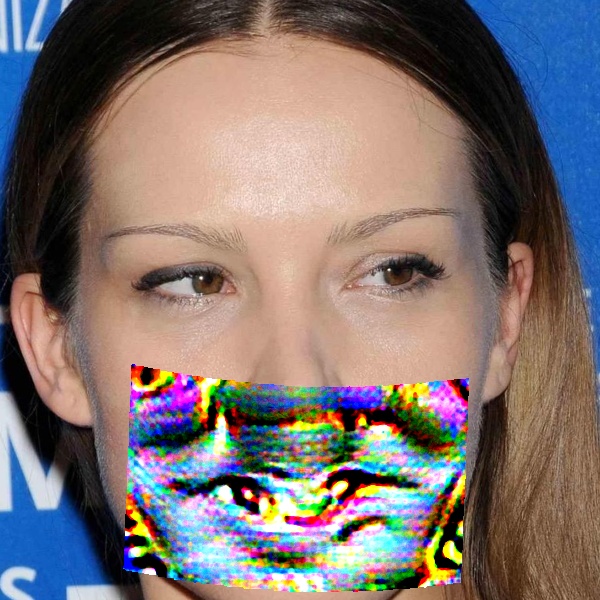} &
    \includegraphics[width=0.18\linewidth]{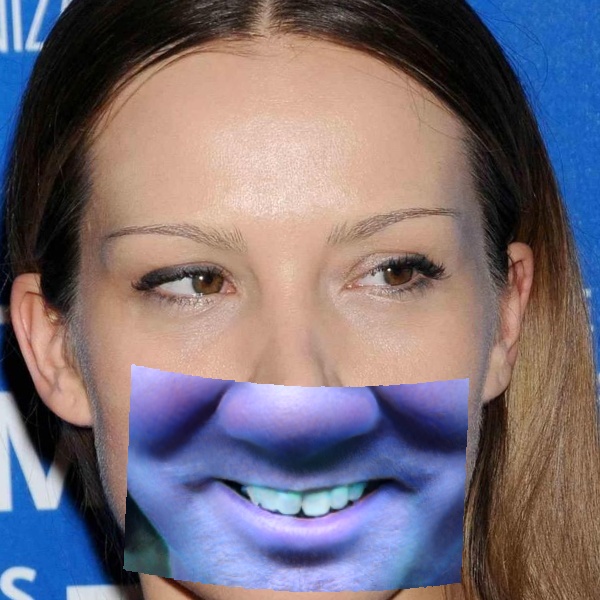}
    \\
    \includegraphics[width=0.18\linewidth]{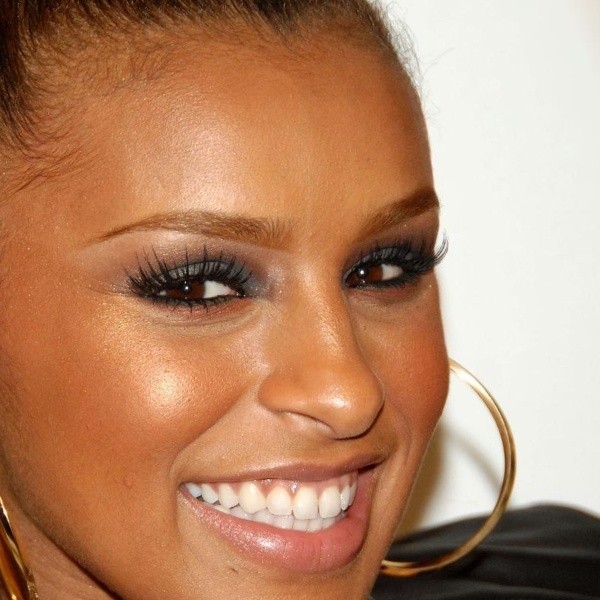} &
    \includegraphics[width=0.18\linewidth]{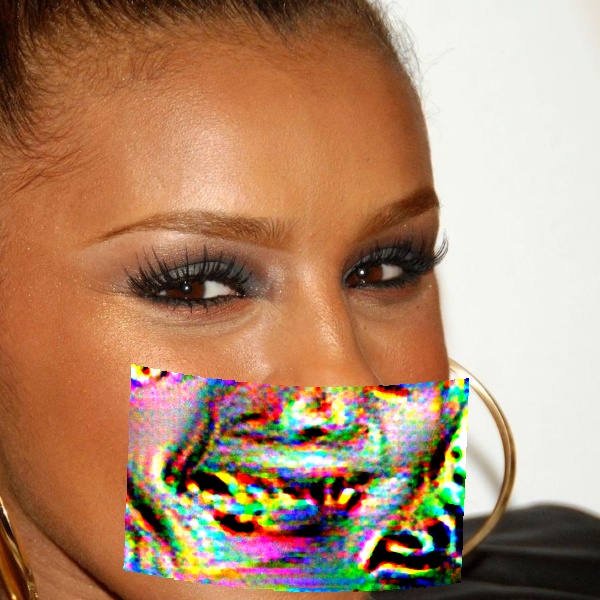} &
    \includegraphics[width=0.18\linewidth]{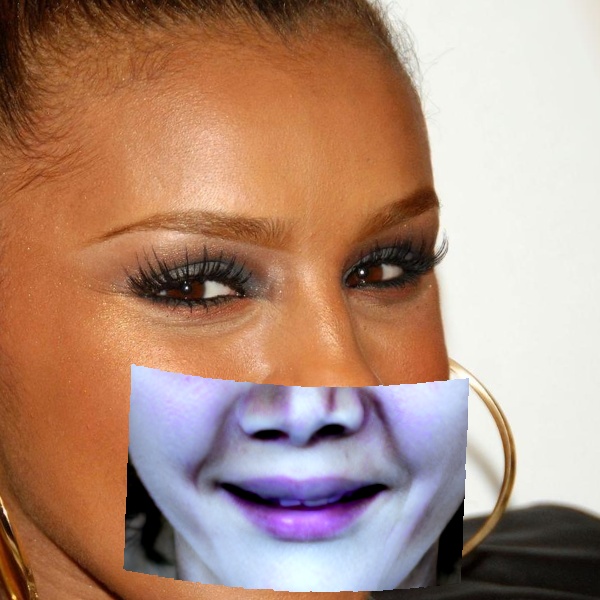}
    \\
    \includegraphics[width=0.18\linewidth]{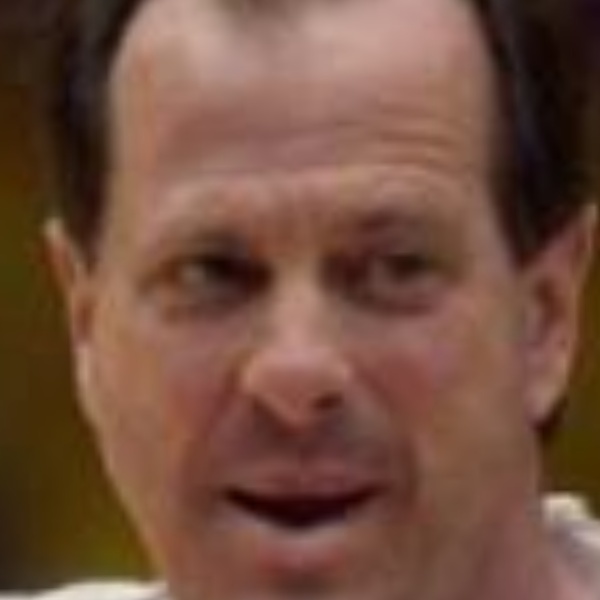} &
    \includegraphics[width=0.18\linewidth]{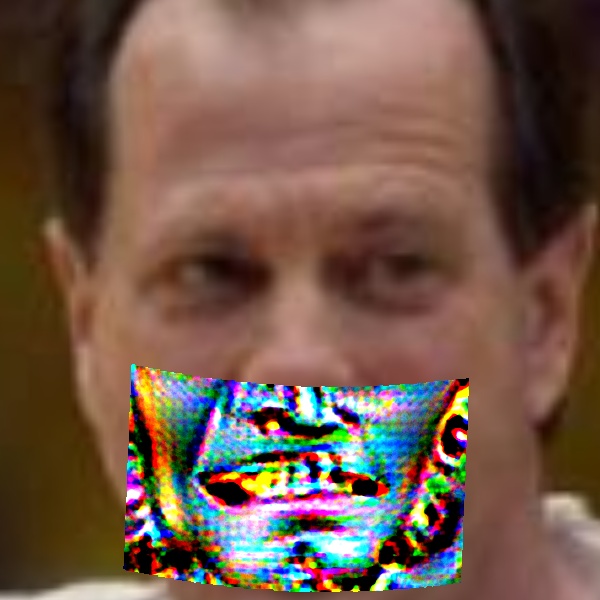} &
    \includegraphics[width=0.18\linewidth]{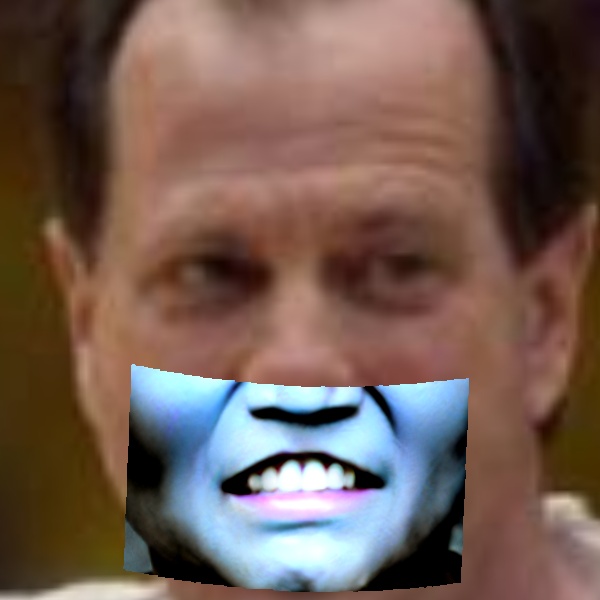}
    \\
    \includegraphics[width=0.18\linewidth]{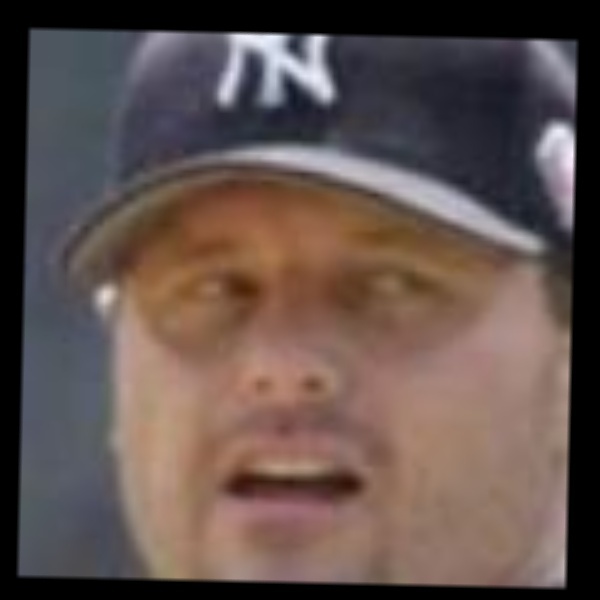} &
    \includegraphics[width=0.18\linewidth]{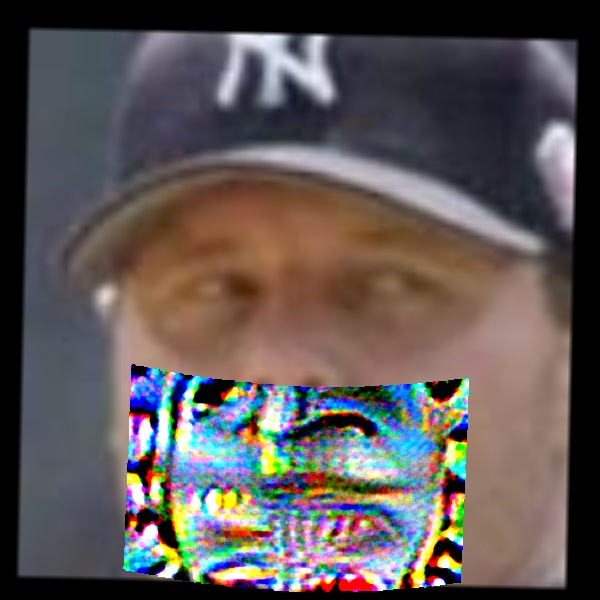} &
    \includegraphics[width=0.18\linewidth]{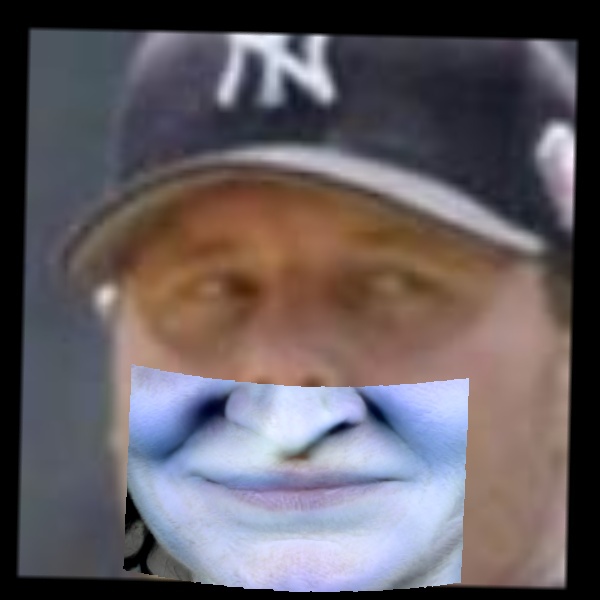}
    \\
    \includegraphics[width=0.18\linewidth]{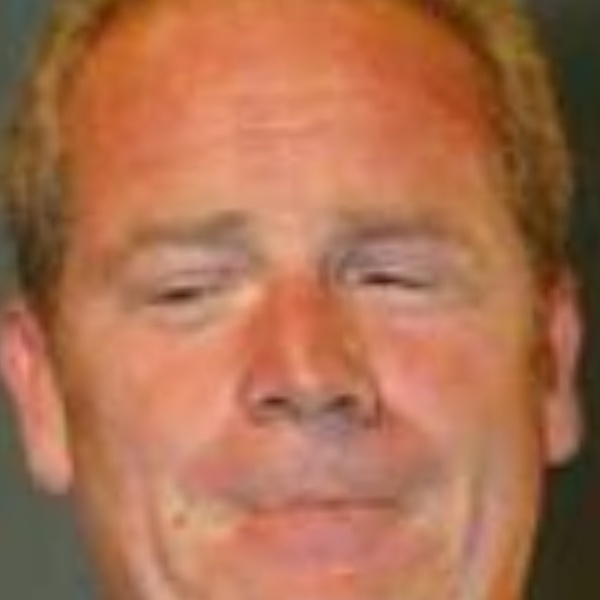} &
    \includegraphics[width=0.18\linewidth]{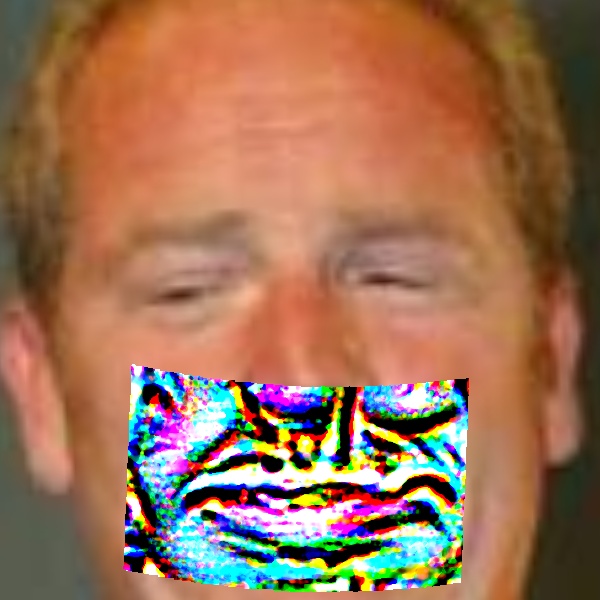} &
    \includegraphics[width=0.18\linewidth]{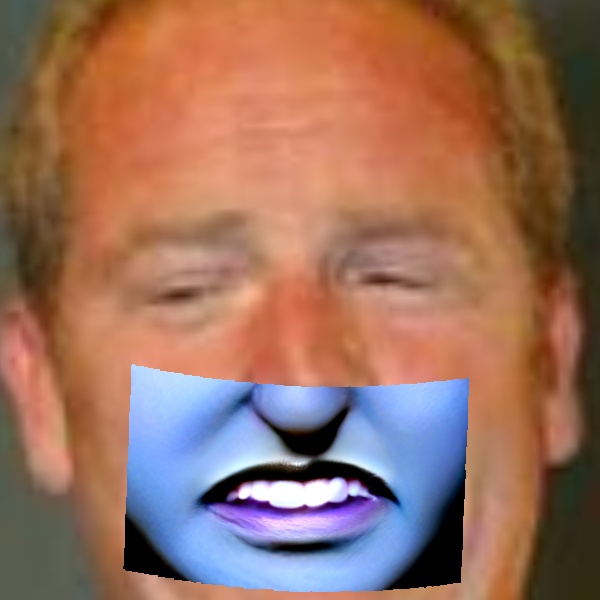}
    \\
    Attacker & TAP-TIDIM & GenAP-DI \\
\end{tabular}
\caption{Visualization of adversarial respirators generated by the TAP-TIDIM and the GenAP-DI methods for dodging attack. The first three rows are the demonstrations on CelebA-HQ dataset and the others are from LFW dataset. And the three columns denotes the pictures of attackers and attackers with the adversarial respirators by generated by TAP-TIDIM and GenAP-DI methods separately. In TAP-TIDIM, we use $\epsilon=255$.}
\label{fig:dodging-respirator} \vspace{-0.5em}
\end{figure*}

\begin{figure*}[htp]
\centering
\begin{tabular}{cccc}
    \includegraphics[width=0.18\linewidth]{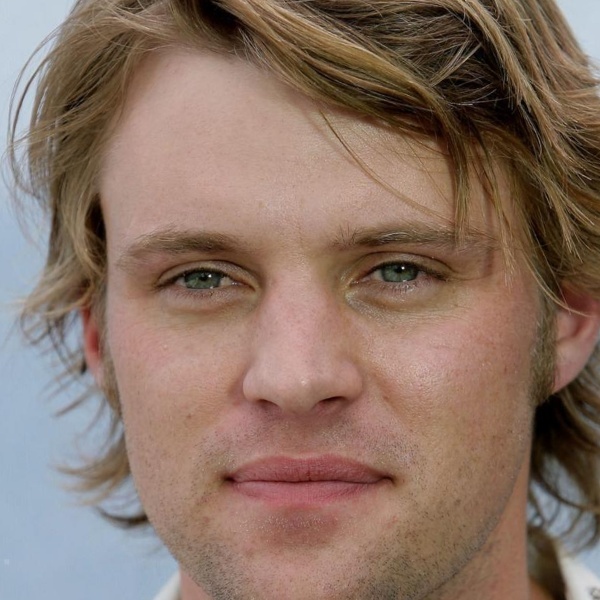} &
    \includegraphics[width=0.18\linewidth]{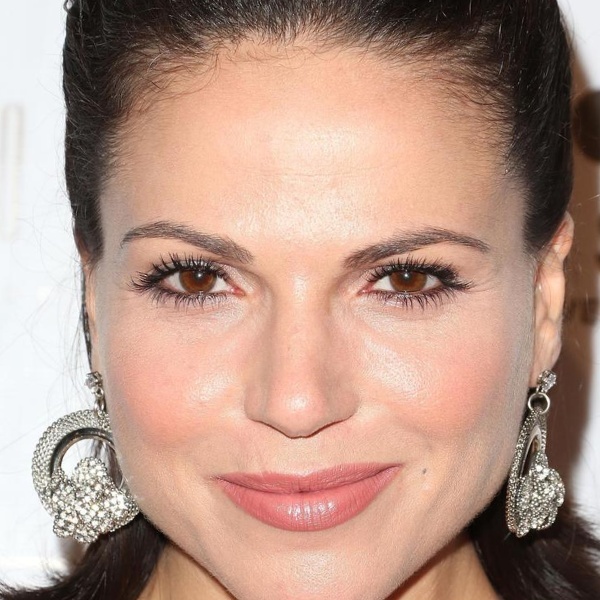} &
    \includegraphics[width=0.18\linewidth]{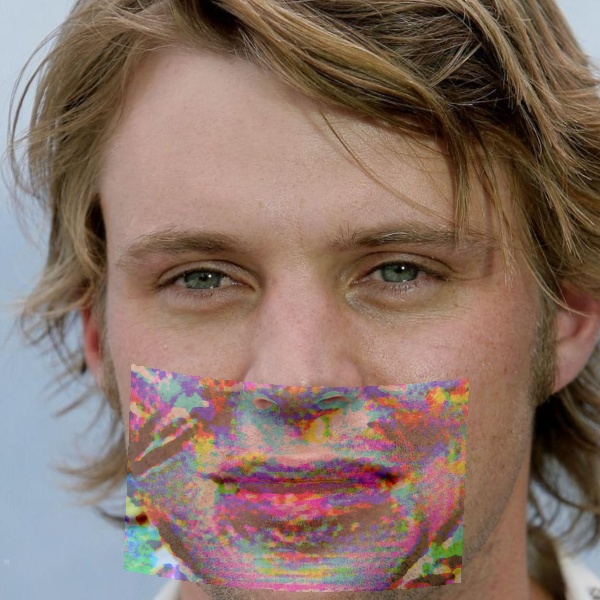} &
    \includegraphics[width=0.18\linewidth]{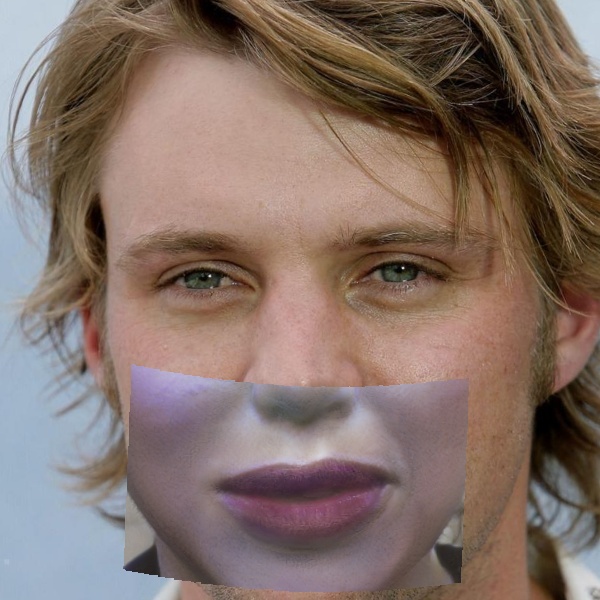}
    \\
    \includegraphics[width=0.18\linewidth]{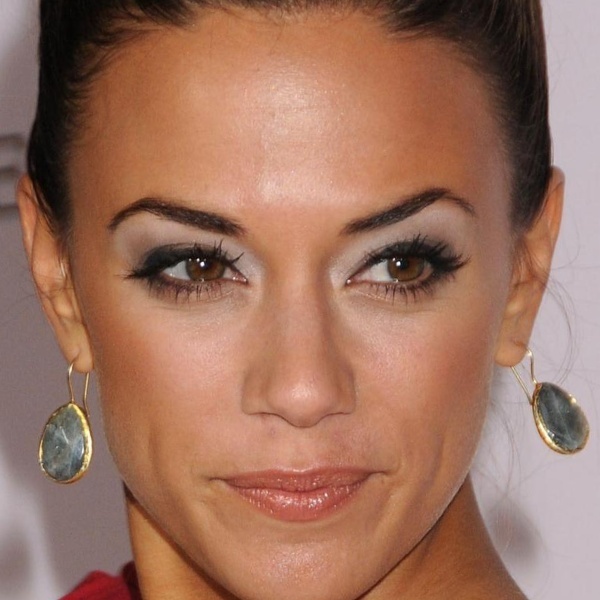} &
    \includegraphics[width=0.18\linewidth]{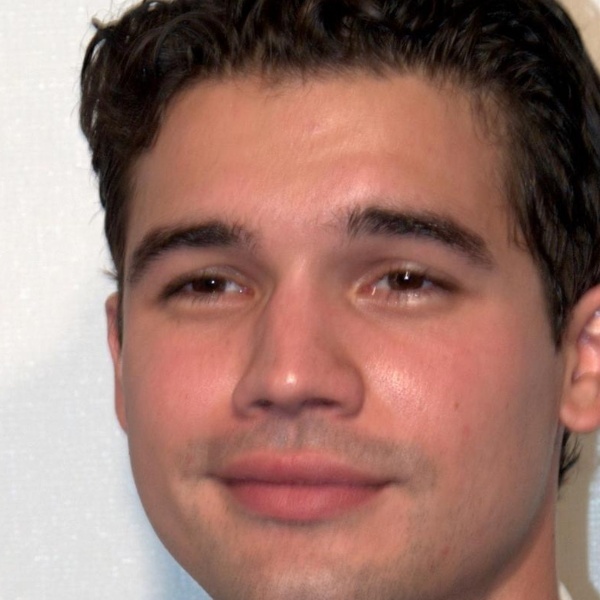} &
    \includegraphics[width=0.18\linewidth]{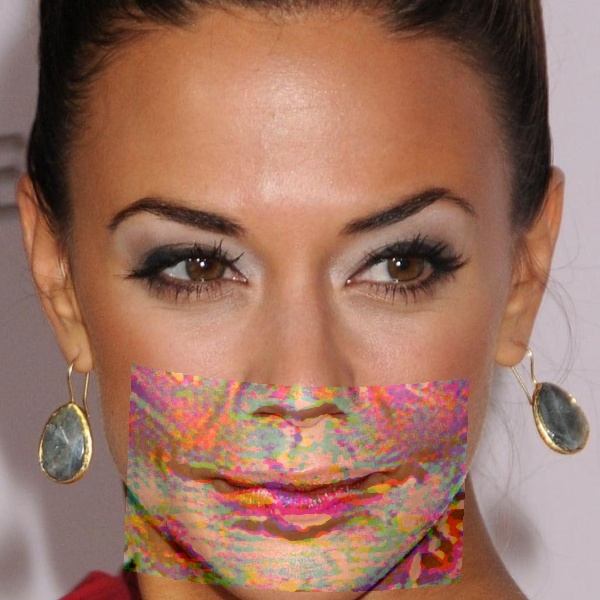} &
    \includegraphics[width=0.18\linewidth]{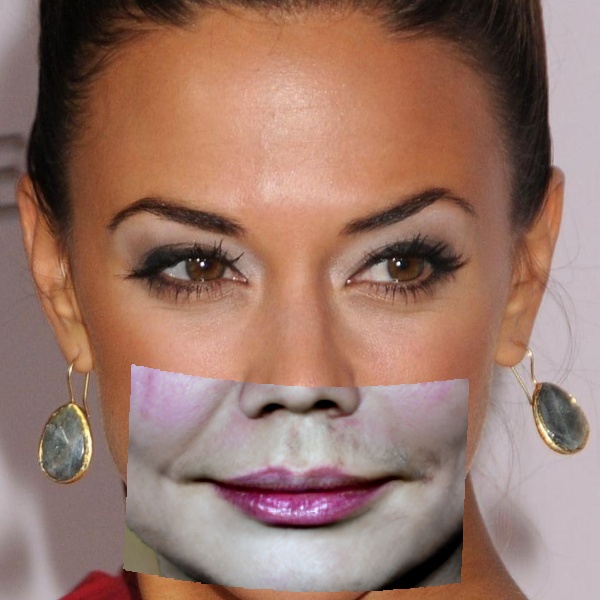}
    \\
    \includegraphics[width=0.18\linewidth]{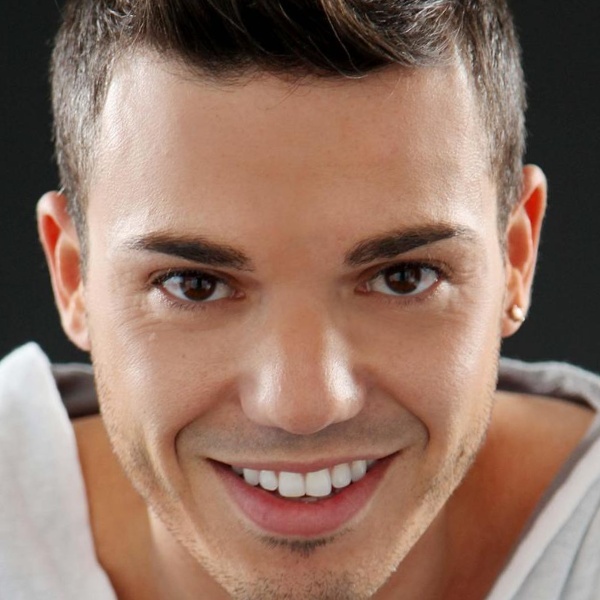} &
    \includegraphics[width=0.18\linewidth]{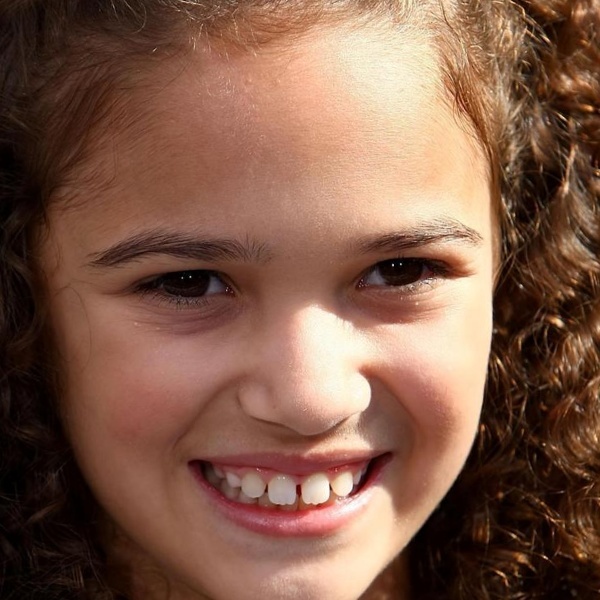} &
    \includegraphics[width=0.18\linewidth]{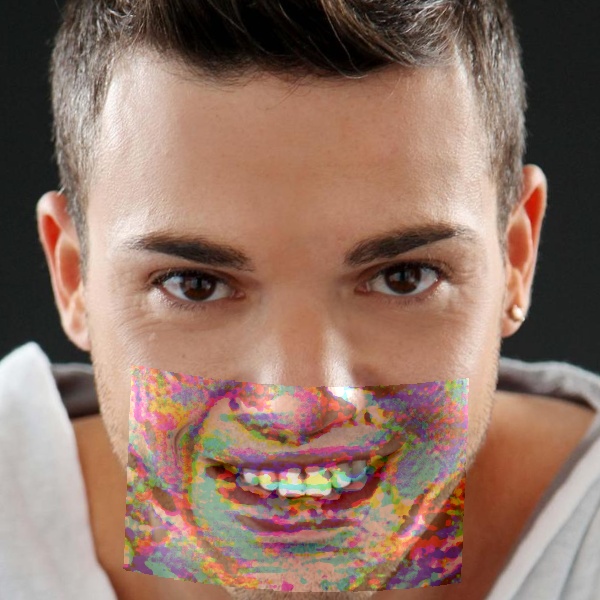} &
    \includegraphics[width=0.18\linewidth]{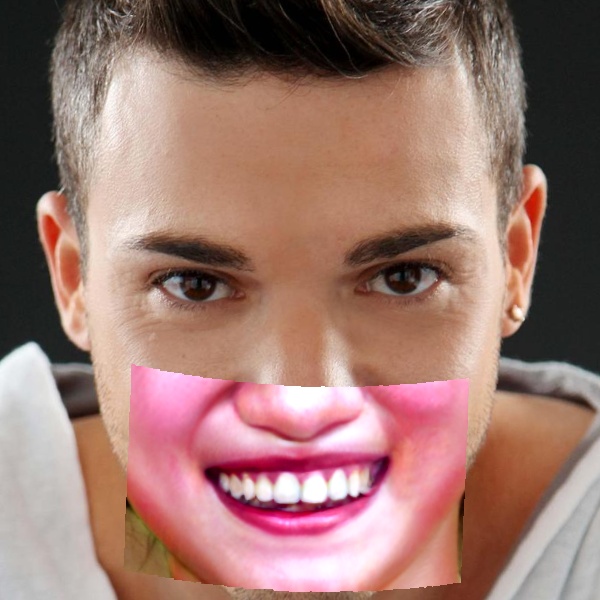}
    \\
    \includegraphics[width=0.18\linewidth]{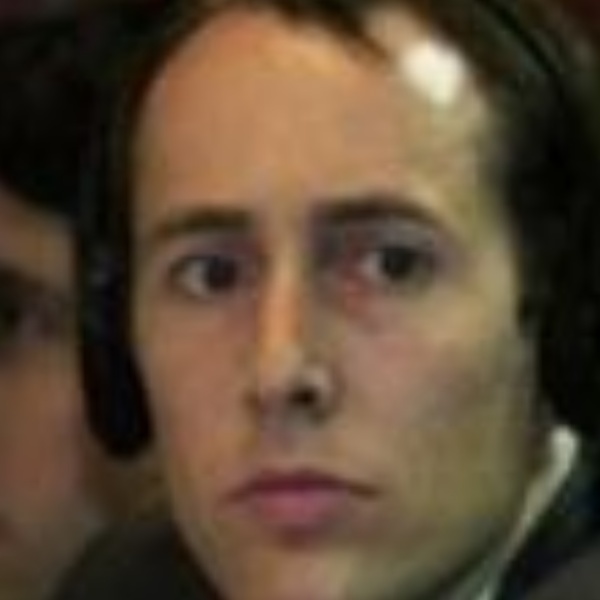} &
    \includegraphics[width=0.18\linewidth]{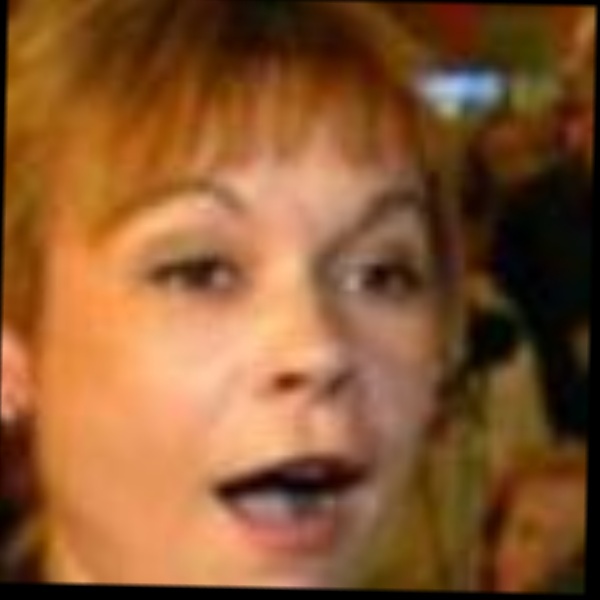} &
    \includegraphics[width=0.18\linewidth]{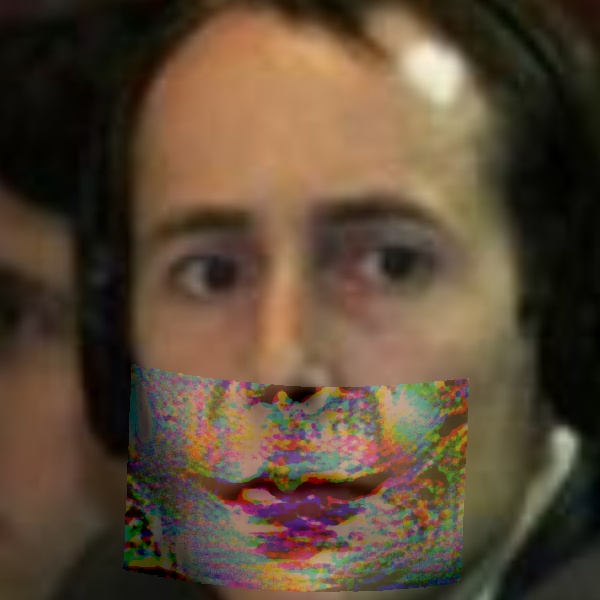} &
    \includegraphics[width=0.18\linewidth]{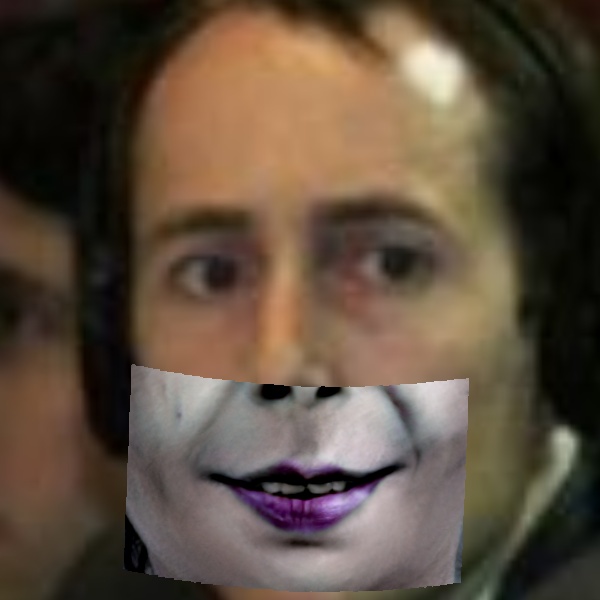}
    \\
    \includegraphics[width=0.18\linewidth]{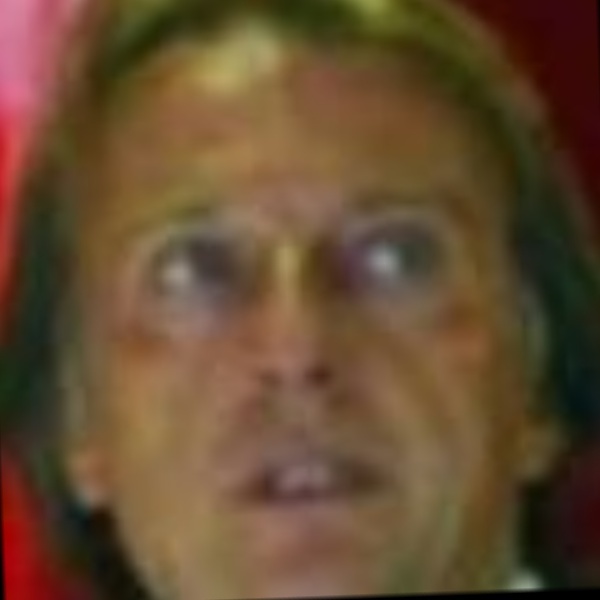} &
    \includegraphics[width=0.18\linewidth]{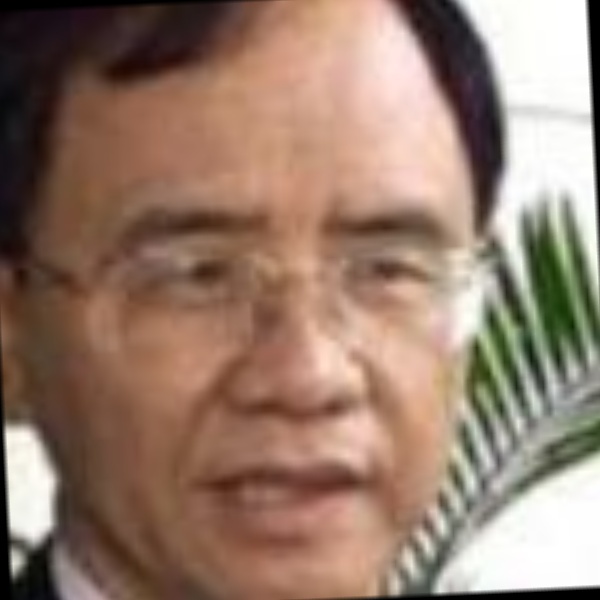} &
    \includegraphics[width=0.18\linewidth]{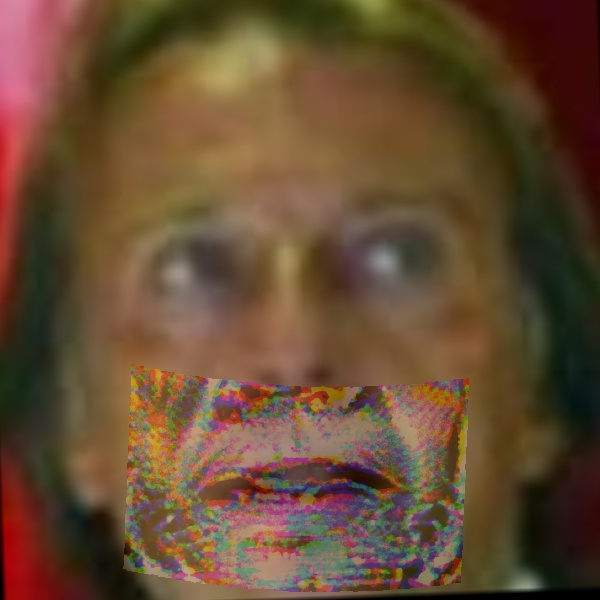} &
    \includegraphics[width=0.18\linewidth]{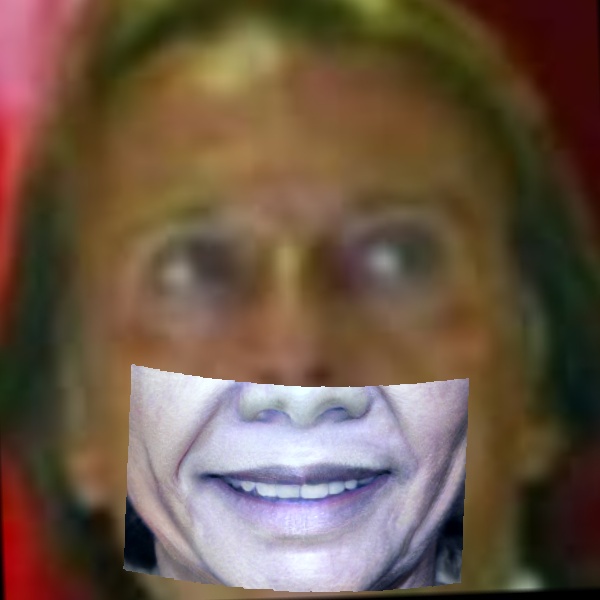}
    \\
    \includegraphics[width=0.18\linewidth]{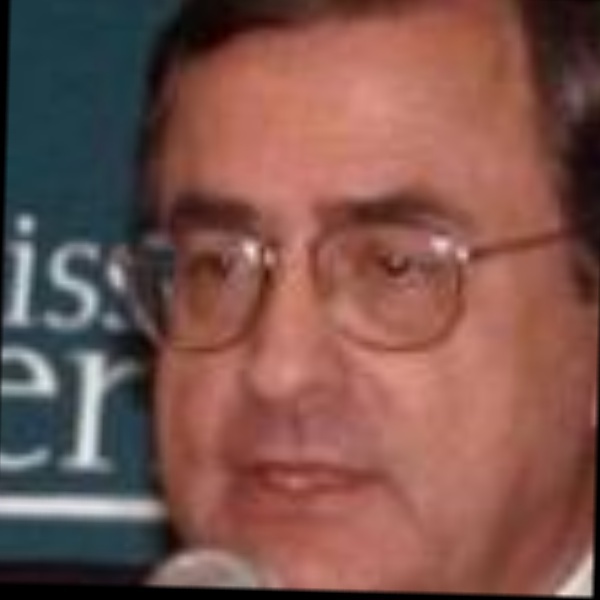} &
    \includegraphics[width=0.18\linewidth]{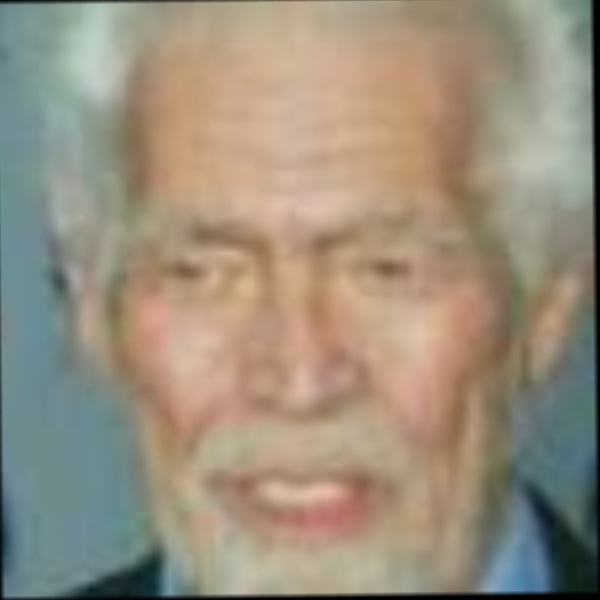} &
    \includegraphics[width=0.18\linewidth]{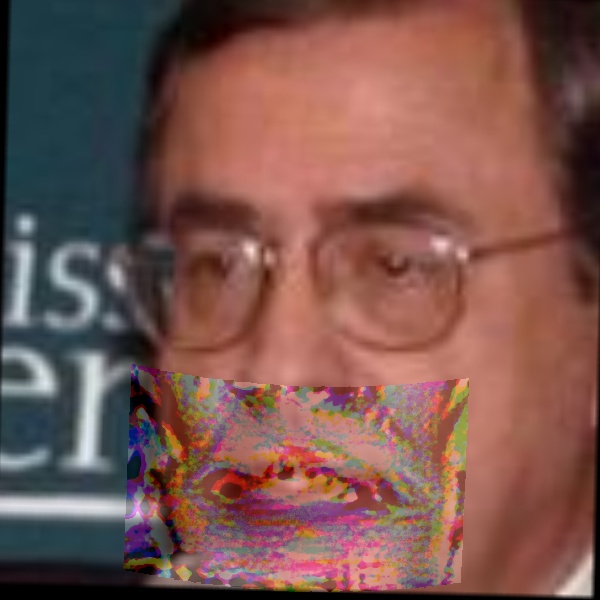} &
    \includegraphics[width=0.18\linewidth]{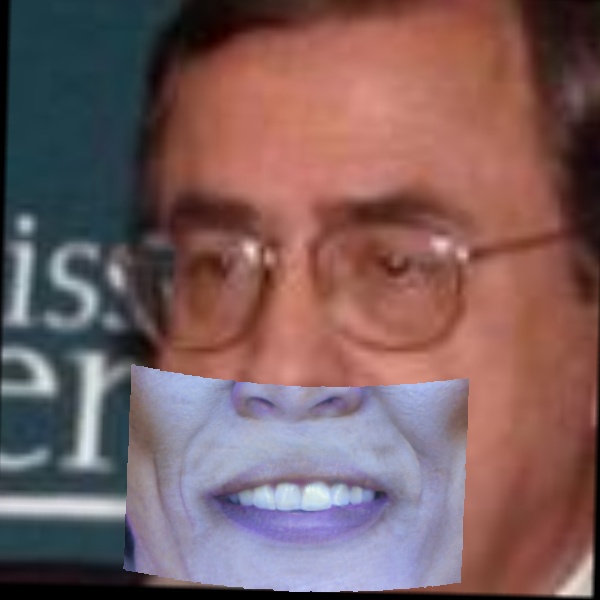}
    \\
    Attacker & Target identity & TAP-TIDIM & GenAP-DI \\
\end{tabular}
\caption{Visualization of adversarial respirators generated by the TAP-TIDIM and the GenAP-DI methods for impersonation attack. Table format follows the Fig.\ref{fig:impersonation-eyeglass}. The first three rows are the demonstrations on CelebA-HQ dataset and the others are from LFW dataset. The first two columns are the photos of attackers and their target identities, and the following two columns show the attackers with the adversarial respirators generated by the proposed TAP-TIDIM and GenAP-DI methods. In TAP-TIDIM, we use $\epsilon=40$.}
\label{fig:impersonation-respirator} \vspace{-0.5em}
\end{figure*}

In this section, we present the results on adversarial respirators to show the generalization of the proposed methods to different regions of the adversarial patches. Tab.~\ref{tab:dodging-verification-mask},~\ref{tab:impersonate-verification-mask} show the results on dodging and impersonation attack, respectively, on the face verification task. Tab.~\ref{tab:dodging-identification-respirator}, \ref{tab:impersonate-identification-respirator} show the results on face identification task. The conclusions are consistent with those drawn from the adversarial eyeglass frames. We visualize the adversarial examples in Fig.~\ref{fig:dodging-respirator} and Fig.~\ref{fig:impersonation-respirator}.

\begin{table*}[t]
    \begin{center}
    \small
    \newcommand{\tabincell}[2]{\begin{tabular}{@{}#1@{}}#2\end{tabular}}

    \begin{tabular}{c|c|c|c|c|c|c|c}
    \hline
        \multirow{2}{*}{} & \multirow{2}{*}{Attack} & \multicolumn{3}{c|}{CelebA-HQ} & \multicolumn{3}{c}{LFW} \\
         \cline{3-8}
         &  & ArcFace & CosFace & FaceNet & ArcFace & CosFace & FaceNet \\
         \hline
         ArcFace& \tabincell{c}{TAP-MIM \\ TAP-TIDIM \\ GenAP \\ GenAP-DI} &
        \tabincell{c}{$\mathbf{1.0000}^*$\\$\mathbf{1.0000}^*$\\$\mathbf{1.0000}^*$\\$\mathbf{1.0000}^*$\\} &
        \tabincell{c}{$0.0800$\\$0.1175$\\$\mathbf{0.3875}$\\$0.3150$\\} &
        \tabincell{c}{$0.3050$\\$0.3300$\\$\mathbf{0.8875}$\\$0.8150$\\} &
        \tabincell{c}{$0.9950^*$\\$\mathbf{1.0000}^*$\\$\mathbf{1.0000}^*$\\$\mathbf{1.0000}^*$\\} &
        \tabincell{c}{$0.0600$\\$0.0875$\\$\mathbf{0.2950}$\\$0.2675$\\} &
        \tabincell{c}{$0.1175$\\$0.1275$\\$\mathbf{0.7300}$\\$0.6325$\\}
        \\
         \hline
         CosFace& \tabincell{c}{TAP-MIM \\ TAP-TIDIM \\ GenAP \\ GenAP-DI} &
        \tabincell{c}{$0.0400$\\$0.0650$\\$0.5425$\\$\mathbf{0.5725}$\\} &
        \tabincell{c}{$0.9975^*$\\$0.9975^*$\\$\mathbf{1.0000}^*$\\$\mathbf{1.0000}^*$\\} &
        \tabincell{c}{$0.2600$\\$0.2825$\\$\mathbf{0.8650}$\\$\mathbf{0.8650}$\\} &
        \tabincell{c}{$0.0250$\\$0.0200$\\$0.3975$\\$\mathbf{0.4025}$\\} &
        \tabincell{c}{$0.9975^*$\\$\mathbf{1.0000}^*$\\$\mathbf{1.0000}^*$\\$\mathbf{1.0000}^*$\\} &
        \tabincell{c}{$0.0975$\\$0.1150$\\$0.7350$\\$\mathbf{0.7750}$\\}
         \\
         \hline
         FaceNet& \tabincell{c}{TAP-MIM \\ TAP-TIDIM \\ GenAP \\ GenAP-DI} &
        \tabincell{c}{$0.0475$\\$0.0425$\\$\mathbf{0.2200}$\\$0.1925$\\} &
        \tabincell{c}{$0.0525$\\$0.0425$\\$\mathbf{0.1375}$\\$\mathbf{0.1375}$\\} &
        \tabincell{c}{$0.9675^*$\\$0.9925^*$\\$0.9950^*$\\$\mathbf{0.9975}^*$\\} &
        \tabincell{c}{$0.0125$\\$0.0100$\\$0.1425$\\$\mathbf{0.1500}$\\} &
        \tabincell{c}{$0.0575$\\$0.0375$\\$\mathbf{0.1400}$\\$\mathbf{0.1400}$\\} &
        \tabincell{c}{$0.9425^*$\\$0.9800^*$\\$0.9850^*$\\$\mathbf{0.9900}^*$\\}
        \\
         \hline
         \hline
    \end{tabular}
    \end{center}
    \caption{The success rates of black-box dodging attack on FaceNet, CosFace, ArcFace in the digital world under the face verification task. The adversarial examples are generated against FaceNet, CosFace, and ArcFace by restricting the adversarial patches to a respirator region. $^*$ indicates white-box attacks.}
    \label{tab:dodging-verification-mask}
\end{table*}

\begin{table*}[t]
    \begin{center}
    \small
    \newcommand{\tabincell}[2]{\begin{tabular}{@{}#1@{}}#2\end{tabular}}

    \begin{tabular}{c|c|c|c|c|c|c|c}
    \hline
        \multirow{2}{*}{} & \multirow{2}{*}{Attack} & \multicolumn{3}{c|}{CelebA-HQ} & \multicolumn{3}{c}{LFW} \\
         \cline{3-8}
         &  & ArcFace & CosFace & FaceNet & ArcFace & CosFace & FaceNet \\
         \hline
         ArcFace& \tabincell{c}{TAP-MIM \\ TAP-TIDIM \\ TAP-TIDIMv2 \\ GenAP \\ GenAP-DI} &
        \tabincell{c}{$\mathbf{1.0000}^*$\\$\mathbf{1.0000}^*$\\$\mathbf{1.0000}^*$\\$\mathbf{1.0000}^*$\\$\mathbf{1.0000}^*$\\} &
        \tabincell{c}{$0.2625$\\$0.3125$\\$0.4250$\\$\mathbf{0.4850}$\\$0.4450$\\} &
        \tabincell{c}{$0.1225$\\$0.1450$\\$0.1975$\\$\mathbf{0.3025}$\\$0.2800$\\} &
        \tabincell{c}{$0.9850^*$\\$\mathbf{1.0000}^*$\\$\mathbf{1.0000}^*$\\$0.9975^*$\\$\mathbf{1.0000}^*$\\} &
        \tabincell{c}{$0.2450$\\$0.2800$\\$0.3725$\\$\mathbf{0.4675}$\\$0.4525$\\} &
        \tabincell{c}{$0.1025$\\$0.1125$\\$0.1475$\\$\mathbf{0.2750}$\\$0.2275$\\}
        \\
         \hline
         CosFace& \tabincell{c}{TAP-MIM \\ TAP-TIDIM \\ TAP-TIDIMv2 \\ GenAP \\ GenAP-DI} &
        \tabincell{c}{$0.4100$\\$0.4025$\\$0.5425$\\$0.6250$\\$\mathbf{0.6325}$\\} &
        \tabincell{c}{$\mathbf{1.0000}^*$\\$\mathbf{1.0000}^*$\\$\mathbf{1.0000}^*$\\$\mathbf{1.0000}^*$\\$\mathbf{1.0000}^*$\\} &
        \tabincell{c}{$0.1475$\\$0.1750$\\$0.2250$\\$\mathbf{0.2975}$\\$\mathbf{0.2975}$\\} &
        \tabincell{c}{$0.2500$\\$0.2225$\\$0.3400$\\$0.5025$\\$\mathbf{0.5050}$\\} &
        \tabincell{c}{$0.9925^*$\\$\mathbf{1.0000}^*$\\$\mathbf{1.0000}^*$\\$\mathbf{1.0000}^*$\\$0.9975^*$\\} &
        \tabincell{c}{$0.1200$\\$0.1300$\\$0.1775$\\$0.2900$\\$\mathbf{0.3100}$\\}
         \\
         \hline
         FaceNet& \tabincell{c}{TAP-MIM \\ TAP-TIDIM \\ TAP-TIDIMv2 \\ GenAP \\ GenAP-DI} &
        \tabincell{c}{$0.1925$\\$0.1900$\\$\mathbf{0.3700}$\\$0.1950$\\$0.2025$\\} &
        \tabincell{c}{$0.1625$\\$0.1950$\\$\mathbf{0.2700}$\\$0.1525$\\$0.1550$\\} &
        \tabincell{c}{$0.6525^*$\\$\mathbf{0.8700}^*$\\$0.8600^*$\\$0.7075^*$\\$0.7325^*$\\} &
        \tabincell{c}{$0.0900$\\$0.1000$\\$\mathbf{0.2275}$\\$0.1475$\\$0.1325$\\} &
        \tabincell{c}{$0.1450$\\$0.1725$\\$\mathbf{0.2400}$\\$0.1800$\\$0.1700$\\} &
        \tabincell{c}{$0.6325^*$\\$0.8550^*$\\$\mathbf{0.8575}^*$\\$0.7550^*$\\$0.7450^*$\\}
        \\
         \hline
         \hline
    \end{tabular}
    \end{center}
    \caption{The success rates of black-box impersonation attack on FaceNet, CosFace, ArcFace in the digital world under the face verification task. The adversarial examples are generated against FaceNet, CosFace, and ArcFace by restricting the adversarial patches to a respirator region. $^*$ indicates white-box attacks.}
    \label{tab:impersonate-verification-mask}
\end{table*}

\begin{table*}[t]
    \begin{center}
    \footnotesize
    \newcommand{\tabincell}[2]{\begin{tabular}{@{}#1@{}}#2\end{tabular}}

    \begin{tabular}{c|c|c|c|c|c|c|c}
    \hline
        \multirow{2}{*}{} & \multirow{2}{*}{Attack} & \multicolumn{3}{c|}{CelebA-HQ} & \multicolumn{3}{c}{LFW} \\
         \cline{3-8}
         &  & ArcFace & CosFace & FaceNet & ArcFace & CosFace & FaceNet \\
         \hline
         ArcFace& \tabincell{c}{TAP-MIM \\ TAP-TIDIM \\ GenAP \\ GenAP-DI} &
        \tabincell{c}{$\mathbf{1.0000}^*$\\$\mathbf{1.0000}^*$\\$\mathbf{1.0000}^*$\\$\mathbf{1.0000}^*$\\} &
        \tabincell{c}{$0.1875$\\$0.2350$\\$\mathbf{0.5450}$\\$0.4850$\\} &
        \tabincell{c}{$0.4800$\\$0.4925$\\$\mathbf{0.9400}$\\$0.9000$\\} &
        \tabincell{c}{$\mathbf{1.0000}^*$\\$\mathbf{1.0000}^*$\\$\mathbf{1.0000}^*$\\$\mathbf{1.0000}^*$\\} &
        \tabincell{c}{$0.1075$\\$0.1550$\\$\mathbf{0.5025}$\\$0.4400$\\} &
        \tabincell{c}{$0.2375$\\$0.2550$\\$\mathbf{0.8525}$\\$0.7950$\\} \\
         \hline
         CosFace& \tabincell{c}{TAP-MIM \\ TAP-TIDIM \\ GenAP \\ GenAP-DI} &
        \tabincell{c}{$0.1250$\\$0.1650$\\$\mathbf{0.7325}$\\$\mathbf{0.7325}$\\} &
        \tabincell{c}{$0.9975^*$\\$0.9975^*$\\$0.9975^*$\\$\mathbf{1.0000}^*$\\} &
        \tabincell{c}{$0.4500$\\$0.4500$\\$\mathbf{0.9250}$\\$0.9150$\\} &
        \tabincell{c}{$0.0325$\\$0.0300$\\$\mathbf{0.5750}$\\$0.5500$\\} &
        \tabincell{c}{$\mathbf{1.0000}^*$\\$\mathbf{1.0000}^*$\\$\mathbf{1.0000}^*$\\$\mathbf{1.0000}^*$\\} &
        \tabincell{c}{$0.2075$\\$0.2375$\\$0.8850$\\$\mathbf{0.9075}$\\} \\
         \hline
         FaceNet& \tabincell{c}{TAP-MIM \\ TAP-TIDIM \\ GenAP \\ GenAP-DI} &
        \tabincell{c}{$0.1300$\\$0.0900$\\$\mathbf{0.3500}$\\$0.3350$\\} &
        \tabincell{c}{$0.1475$\\$0.1325$\\$0.2725$\\$\mathbf{0.2800}$\\} &
        \tabincell{c}{$0.9800^*$\\$0.9950^*$\\$\mathbf{0.9975}^*$\\$\mathbf{0.9975}^*$\\} &
        \tabincell{c}{$0.0425$\\$0.0200$\\$0.2425$\\$\mathbf{0.2600}$\\} &
        \tabincell{c}{$0.1200$\\$0.0750$\\$0.3275$\\$\mathbf{0.3425}$\\} &
        \tabincell{c}{$0.9650^*$\\$0.9850^*$\\$0.9875^*$\\$\mathbf{0.9950}^*$\\} \\
         \hline
         \hline
    \end{tabular}
    \end{center}
    \caption{The success rates of black-box dodging attack on FaceNet, CosFace, ArcFace in the digital world under the face identification task. The adversarial examples are generated against FaceNet, CosFace, and ArcFace by restricting the adversarial patches to a respirator frame region. $^*$ indicates white-box attacks.}
    \label{tab:dodging-identification-respirator}
\end{table*}

\begin{table*}[t]
    \begin{center}
    \small
    \newcommand{\tabincell}[2]{\begin{tabular}{@{}#1@{}}#2\end{tabular}}

    \begin{tabular}{c|c|c|c|c|c|c|c}
    \hline
        \multirow{2}{*}{} & \multirow{2}{*}{Attack} & \multicolumn{3}{c|}{CelebA-HQ} & \multicolumn{3}{c}{LFW} \\
         \cline{3-8}
         &  & ArcFace & CosFace & FaceNet & ArcFace & CosFace & FaceNet\\
         \hline
         ArcFace& \tabincell{c}{TAP-MIM \\ TAP-TIDIM \\ TAP-TIDIMv2 \\ GenAP \\ GenAP-DI} &
\tabincell{c}{$0.6650^*$\\$0.9350^*$\\$\mathbf{0.9625}^*$\\$0.8800^*$\\$0.9525^*$\\} &
\tabincell{c}{$0.0575$\\$0.0750$\\$0.1875$\\$\mathbf{0.2425}$\\$0.1850$\\} &
\tabincell{c}{$0.0200$\\$0.0275$\\$0.0675$\\$\mathbf{0.1325}$\\$0.0975$\\} &
\tabincell{c}{$0.6555^*$\\$0.9425^*$\\$\mathbf{0.9650}^*$\\$0.8800^*$\\$0.9625^*$\\} &
\tabincell{c}{$0.0600$\\$0.0875$\\$0.1500$\\$0.2025$\\$\mathbf{0.2075}$\\} &
\tabincell{c}{$0.0200$\\$0.0300$\\$0.0600$\\$\mathbf{0.1275}$\\$0.0975$\\} \\
         \hline
         CosFace& \tabincell{c}{TAP-MIM \\ TAP-TIDIM \\ TAP-TIDIMv2 \\ GenAP \\ GenAP-DI} &
\tabincell{c}{$0.1100$\\$0.1150$\\$0.2125$\\$\mathbf{0.2725}$\\$0.2550$\\} &
\tabincell{c}{$0.6900^*$\\$0.8875^*$\\$\mathbf{0.9375}^*$\\$0.8725^*$\\$0.8825^*$\\} &
\tabincell{c}{$0.0375$\\$0.0375$\\$0.0825$\\$0.1275$\\$\mathbf{0.1300}$\\} &
\tabincell{c}{$0.0800$\\$0.0950$\\$0.1425$\\$0.2200$\\$\mathbf{0.2275}$\\} &
\tabincell{c}{$0.7125^*$\\$0.9425^*$\\$\mathbf{0.9875}^*$\\$0.9325^*$\\$0.9225^*$\\} &
\tabincell{c}{$0.0450$\\$0.0450$\\$0.0800$\\$0.1375$\\$\mathbf{0.1400}$\\} \\
         \hline
         FaceNet& \tabincell{c}{TAP-MIM \\ TAP-TIDIM \\ TAP-TIDIMv2 \\ GenAP \\ GenAP-DI} &
\tabincell{c}{$0.0425$\\$0.0475$\\$\mathbf{0.1275}$\\$0.0450$\\$0.0525$\\} &
\tabincell{c}{$0.0325$\\$0.0550$\\$\mathbf{0.0975}$\\$0.0550$\\$0.0400$\\} &
\tabincell{c}{$0.3250^*$\\$0.6375^*$\\$\mathbf{0.6800}^*$\\$0.4400^*$\\$0.4700^*$\\} &
\tabincell{c}{$0.0350$\\$0.0300$\\$\mathbf{0.0900}$\\$0.0625$\\$0.0575$\\} &
\tabincell{c}{$0.0275$\\$0.0425$\\$\mathbf{0.0850}$\\$0.0725$\\$0.0575$\\} &
\tabincell{c}{$0.3000^*$\\$0.5825^*$\\$\mathbf{0.6550}^*$\\$0.5100^*$\\$0.4950^*$\\} \\
         \hline
         \hline
    \end{tabular}
    \end{center}
    \caption{The success rates of black-box impersonation attack on FaceNet, CosFace, ArcFace in the digital world under the face identification task. The adversarial examples are generated against FaceNet, CosFace, and ArcFace by restricting the adversarial patches to a respirator frame region. $^*$ indicates white-box attacks.}
    \label{tab:impersonate-identification-respirator}
\end{table*}

\section{Experiments on SemanticAdv}
SemanticAdv~\cite{qiu2020semanticadv} uses StarGAN~\cite{choi2018stargan} to adversarially perturb the attributes in a face image. This is very similar with our idea of using face-like features to generate adversarial perturbations. Nevertheless, SemanticAdv proposed to generate the adversarial example by interpolation between two feature maps. Their algorithm is designed for the imperceptible setting, where the adversarial perturbation should be indistinguiable from the original image. However, the slight perturbation limits their performance in the patch setting, where the perturbation is allowed to be perceptible. Instead, the proposed GenAP algorithms can leverage the relaxation on perceptibility to improve the attack success rates. We perform experiments to verify this point in the following.

Specifically, we use the code from SemanticAdv\footnote{\url{https://github.com/AI-secure/SemanticAdv}} to reproduce their results.  We use the first 100 image pairs from their dataset (Celeba) to compare the algorithms.  The code performs targeted attack on face verification. We modify their code to 1) perturb only the eyeglass frame region, 2) use our substitute models. Since their method generates adversarial examples for each attributes for each attacker-target pair, we consider their attack for an image pair is successful if the attack from any attribute is successful. The results are shown in Tab.~\ref{tab:semanticadv}. Our methods significantly outperforms theirs in the patch setting. In this setting, the slight perturbation generated by their method has difficulty in even white-box attacks.

\begin{table*}[t]
    \begin{center}
    \footnotesize
    \newcommand{\tabincell}[2]{\begin{tabular}{@{}#1@{}}#2\end{tabular}}
    \begin{tabular}{c|c|c|c}
    \hline
         Attack & ArcFace & CosFace & FaceNet  \\
         \hline \hline
         \tabincell{c}{SemanticAdv \\ GenAP} &
          \tabincell{c}{$0.82$ \\ $\mathbf{1.00}$ } &
         \tabincell{c}{$0.19$ \\ $\mathbf{0.51}$ } &
         \tabincell{c}{$0.11$ \\ $\mathbf{0.36}$ } \\
         \hline
    \end{tabular}
    \end{center}
    \caption{The attack success rate of black-box impersonation attack in face verification using SemanticAdv and GenAP on the CelebA dataset. ArcFace is the substitute model.}
    \label{tab:semanticadv}
\end{table*}


\section{Experiments on image classification}
The proposed GenAP methods can be easily extended to the image classification task by replacing the adversarial losses for face recognition (Eq.~(2) in the main text) by the cross-entropy loss~\cite{goodfellow2014explaining} widely used in image classification. This section shows the effectiveness of the GenAP algorithms on the image classification tasks.

We use two datasets, CIFAR10 and ImageNet. The images from these two datasets are $32\times 32$ and $224\times 224$, respectively. The adversarial patch region is designed to be a square at the center of the image. We observe the experimental results by changing the length of the square. The detailed information of the recognition models, generative models and lengths of the square patches are listed in Tab.~\ref{tab:setting}. All models are accessible from the Internet\footnote{CIFAR recognition models: \url{https://github.com/huyvnphan/PyTorch_CIFAR10}, CIFAR10 generative models:  \url{https://github.com/NVlabs/stylegan2-ada-pytorch}, ImageNet recognition models:  \url{https://pytorch.org/vision/stable/models.html}, ImageNet generative models: \url{https://github.com/ajbrock/BigGAN-PyTorch} } To evaluate the attack performance, we report the success rate (higher is better) as the fraction of adversarial images that are misclassified to the desired target class (\ie, targeted attack). For each dataset, we randomly sample $1000$ images. For each image, we randomly sample a distinct class as the target.

For the TAP algorithms, we set $\epsilon=255$.
For the GenAP algorithms on CIFAR10, we introduce several GenAP algorithms that optimize in different latent spaces. StyleGAN2-ADA is a conditional generative model. First, the GenAP-cond algorithm uses the image directly generated by the generative model conditional on the target class.  Second, the GenAP-DI-cond-opt algorithm is the GenAP-DI algorithm optimized in the $\mathbf{Z}$ space, with the conditional variable set to the target class. Third, the GenAP-DI-uncond-opt algorithm is a GenAP-DI algorithm optimized in the $\mathbf{W}^+$ plus the noise space. Similarly, GenAP algorithms are introduced on ImageNet. Since BigGAN is also a conditional generative model, we can also define the corresponding GenAP-cond and GenAP-DI-cond-opt algorithm with it.

The experimental results on CIFAR10 are shown in Tab.~\ref{tab:cifar}. We have the following observations. First, when the patch size is very small, the TAP algorithms achieve higher success rates on white-box attack by leveraging the larger search space (\ie, without regularization). Second, optimizing the latent spaces of the generative models yield better results than naively using the inference results from the conditional generative model (GenAP-DI-uncond-opt and GenAP-DI-cond-opt v.s. GenAP-cond). Third, the GenAP-DI-cond-opt outperforms the TAP-TIDIM in black-box attacks when the patch size is as small as $8\times 8$, which occupies $6\%$ of the whole image.

The experiments results on ImageNet are shown in Tab.~\ref{tab:imagenet}. The observations are consistent with those on CIFAR-10. The GenAP-DI-cond-opt outperforms the TAP-TIDIM in black-box attacks when the patch size is as small as $60\times 60$, which occupies $7\%$ of the whole image.

\begin{table*}[h!]
    \begin{center}
    \footnotesize
    \newcommand{\tabincell}[2]{\begin{tabular}{@{}#1@{}}#2\end{tabular}}
    \begin{tabular}{c|c|c|c}
    \hline
        Dataset & Image classification models & Generative models & Lengths of the square patch \\
         \hline \hline
        CIFAR10 & ResNet50, MobileNetV2, InceptionV3, DenseNet121, VGG16 & StyleGAN2-ADA& $8, 12, 16, 24$\\
        ImageNet & ResNet101, DenseNet121, VGG16, ResNet50 & BigGAN & $40, 80, 60, 80, 100, 120$ \\
         \hline
    \end{tabular}
    \end{center}
    \caption{Setting of the image classification experiments.}
    \label{tab:setting}
\end{table*}

\begin{table*}[t]
    \begin{center}
    \footnotesize
    \newcommand{\tabincell}[2]{\begin{tabular}{@{}#1@{}}#2\end{tabular}}

    \begin{tabular}{c|c|c|c|c|c|c}
    \hline
        Patch size & Attack & ResNet50 & MobileNetV2 & InceptionV3 & DesNet121 & VGG16 \\
         \hline \hline
         $8\times 8$& \tabincell{c}{TAP-TIDIM \\ GenAP-DI-uncond-opt \\ GenAP-cond \\ GenAP-DI-cond-opt} &
          \tabincell{c}{$\mathbf{0.530}^*$ \\ $0.128^*$ \\ $0.037$ \\ $0.462^*$} &
         \tabincell{c}{$0.101$ \\ $0.071$ \\ $0.036$ \\ $\mathbf{0.156}$} &
         \tabincell{c}{$0.114$ \\ $0.092$ \\ $0.040$ \\ $\mathbf{0.185}$} &
         \tabincell{c}{$0.194$ \\ $0.104$ \\ $0.043$ \\ $\mathbf{0.262}$} &
         \tabincell{c}{$0.248$ \\ $0.094$ \\ $0.034$ \\ $\mathbf{0.189}$} \\
         \hline
         $12\times 12$& \tabincell{c}{TAP-TIDIM \\ GenAP-DI-uncond-opt \\ GenAP-cond \\ GenAP-DI-cond-opt} &
          \tabincell{c}{$0.793^*$ \\ $0.246^*$ \\ $0.128$ \\ $\mathbf{0.834}^*$} &
         \tabincell{c}{$0.209$ \\ $0.170$ \\ $0.117$ \\ $\mathbf{0.424}$} &
         \tabincell{c}{$0.224$ \\ $0.198$ \\ $0.131$ \\ $\mathbf{0.458}$} &
         \tabincell{c}{$0.277$ \\ $0.218$ \\ $0.131$ \\ $\mathbf{0.585}$} &
         \tabincell{c}{$0.345$ \\ $0.193$ \\ $0.128$ \\ $\mathbf{0.478}$} \\
         \hline
         $16\times 16$& \tabincell{c}{TAP-TIDIM \\ GenAP-DI-uncond-opt \\ GenAP-cond \\ GenAP-DI-cond-opt} &
          \tabincell{c}{$0.865^*$ \\ $0.333^*$ \\ $0.299$ \\ $\mathbf{0.969}^*$} &
         \tabincell{c}{$0.227$ \\ $0.260$ \\ $0.319$ \\ $\mathbf{0.778}$} &
         \tabincell{c}{$0.289$ \\ $0.254$ \\ $0.335$ \\ $\mathbf{0.778}$} &
         \tabincell{c}{$0.304$ \\ $0.270$ \\ $0.313$ \\ $\mathbf{0.846}$} &
         \tabincell{c}{$0.398$ \\ $0.238$ \\ $0.316$ \\ $\mathbf{0.801}$} \\
         \hline
         $24\times 24$& \tabincell{c}{TAP-TIDIM \\ GenAP-DI-uncond-opt \\ GenAP-cond \\ GenAP-DI-cond-opt} &
          \tabincell{c}{$0.888^*$ \\ $0.461^*$ \\ $0.745$ \\ $\mathbf{1.000}^*$} &
         \tabincell{c}{$0.235$ \\ $0.400$ \\ $0.776$ \\ $\mathbf{0.981}$} &
         \tabincell{c}{$0.325$ \\ $0.398$ \\ $0.808$ \\ $\mathbf{0.989}$} &
         \tabincell{c}{$0.237$ \\ $0.398$ \\ $0.800$ \\ $\mathbf{0.999}$} &
         \tabincell{c}{$0.509$ \\ $0.360$ \\ $0.780$ \\ $\mathbf{0.988}$} \\
         \hline
    \end{tabular}
    \end{center}
    \caption{The success rates of black-box targeted attack on CIFAR-10 dataset using adversarial patches. The adversarial patches are generated against ResNet50.  $^*$ indicates white-box attacks.}
    \label{tab:cifar}
\end{table*}

\begin{table*}[t]
    \begin{center}
    \footnotesize
    \newcommand{\tabincell}[2]{\begin{tabular}{@{}#1@{}}#2\end{tabular}}

    \begin{tabular}{c|c|c|c|c|c}
    \hline
        Patch size & Attack & ResNet101 & DesNet121 & VGG16 & ResNet50\\
         \hline \hline
         $40\times 40$& \tabincell{c}{TAP-TIDIM \\ GenAP-cond \\ GenAP-DI-cond-opt} &
          \tabincell{c}{$\mathbf{0.501}^*$ \\ $0.000$ \\ $0.020^*$ } &
         \tabincell{c}{$\mathbf{0.004}$ \\ $0.001$ \\ $0.000$ } &
         \tabincell{c}{$\mathbf{0.006}$ \\ $0.001$ \\ $0.001$ } &
         \tabincell{c}{$\mathbf{0.006}$ \\ $0.000$ \\ $0.001$ }\\
         \hline
         $60\times 60$& \tabincell{c}{TAP-TIDIM \\ GenAP-cond \\ GenAP-DI-cond-opt} &
          \tabincell{c}{$\mathbf{0.899}^*$ \\ $0.001$ \\ $0.110^*$ } &
         \tabincell{c}{$0.004$ \\ $0.000$ \\ $\mathbf{0.014}$ } &
         \tabincell{c}{$0.001$ \\ $0.000$ \\ $\mathbf{0.009}$ } &
         \tabincell{c}{$0.002$ \\ $0.002$ \\ $\mathbf{0.021}$ }\\
         \hline
         $80\times 80$& \tabincell{c}{TAP-TIDIM \\ GenAP-cond \\ GenAP-DI-cond-opt} &
          \tabincell{c}{$\mathbf{0.993}^*$ \\ $0.012$ \\ $0.011^*$ } &
         \tabincell{c}{$0.002$ \\ $0.005$ \\ $\mathbf{0.037}$ } &
         \tabincell{c}{$0.002$ \\ $0.007$ \\ $\mathbf{0.025}$ } &
         \tabincell{c}{$0.004$ \\ $0.005$ \\ $\mathbf{0.055}$ }\\
         \hline
         $100\times 100$& \tabincell{c}{TAP-TIDIM \\ GenAP-cond \\ GenAP-DI-cond-opt} &
          \tabincell{c}{$\mathbf{1.000}^*$ \\ $0.045$ \\ $0.524^*$ } &
         \tabincell{c}{$0.006$ \\ $0.019$ \\ $\mathbf{0.090}$ } &
         \tabincell{c}{$0.004$ \\ $0.012$ \\ $\mathbf{0.039}$ } &
         \tabincell{c}{$0.002$ \\ $0.025$ \\ $\mathbf{0.129}$ }\\
         \hline
         $120\times 120$& \tabincell{c}{TAP-TIDIM \\ GenAP-cond \\ GenAP-DI-cond-opt} &
          \tabincell{c}{$\mathbf{1.000}^*$ \\ $0.096$ \\ $0.735^*$ } &
         \tabincell{c}{$0.012$ \\ $0.080$ \\ $\mathbf{0.179}$ } &
         \tabincell{c}{$0.006$ \\ $0.039$ \\ $\mathbf{0.066}$ } &
         \tabincell{c}{$0.007$ \\ $0.090$ \\ $\mathbf{0.251}$ }\\
         \hline
    \end{tabular}
    \end{center}
    \caption{The success rates of black-box targeted attack on ImageNet dataset using adversarial patches. The adversarial patches are generated against ResNet101.  $^*$ indicates white-box attacks.}
    \label{tab:imagenet}
\end{table*}

\end{document}